\documentclass[10pt]{article} 
\usepackage[accepted]{tmlr}
\usepackage{hyperref}
\usepackage{url}

\definecolor{darkgray}{gray}{0.3}
\definecolor{BrickRed}{HTML}{B6321C}
\definecolor{MidnightBlue}{HTML}{006795}

\usepackage{amsmath,amsfonts,amssymb,amsthm}
\usepackage{booktabs, caption, makecell}

\usepackage{threeparttable}
\usepackage{progressbar}
\usepackage{wrapfig}

\usepackage{algorithmicx}
\usepackage[ruled]{algorithm}
\usepackage{algpseudocode}

\usepackage{pifont}
\newcommand{\xmark}{\ding{56}}%
\usepackage{chngcntr}
\newtheorem{assumption}{Assumption}

\usepackage{amsfonts} 
\newtheorem{theorem}{Theorem}

\newtheorem{lemma}[theorem]{Lemma}

\usepackage{enumitem}

\newtheorem{corollaryalt}{Corollary}

\newenvironment{corollaryp}[1]{
  \renewcommand\thecorollaryalt{#1}
  \corollaryalt
}{\endcorollaryalt}

\newtheorem{lemmaalt}{Lemma}

\newenvironment{lemmap}[1]{
  \renewcommand\thelemmaalt{#1}
  \lemmaalt
}{\endlemmaalt}

\usepackage{amsfonts}
\usepackage{amsmath}
\usepackage{amsthm}
\usepackage{amssymb}
\usepackage{dsfont}

\usepackage{graphicx}

\usepackage{verbatim}
\usepackage{xspace} 

\usepackage{enumerate}
\usepackage{enumitem}

  \providecommand{\xx}{\mathbf{x}}
  \providecommand{\yy}{\mathbf{y}}

  \usepackage{bm}

\RequirePackage[colorinlistoftodos,bordercolor=orange,backgroundcolor=orange!20,linecolor=orange,textsize=scriptsize]{todonotes}
\providecommand{\mycomment}[3]{\todo[caption={},color=#3!20,inline]{\textbf{#1: }#2}}%
\providecommand{\myinlinecomment}[3]{%
  {\color{#1}#2: #3}}%
\newcommand\commenter[2]%
{%
  \expandafter\newcommand\csname i#1\endcsname[1]{\myinlinecomment{#2}{#1}{##1}}
  \expandafter\newcommand\csname #1\endcsname[1]{\mycomment{#1}{##1}{#2}}
}
  
\usepackage{url}

\renewcommand{\epsilon}{\varepsilon}

\commenter{seb}{red} 
\commenter{yasin}{blue} 
\commenter{bo}{green} 

\title{Synthetic data shuffling accelerates the convergence of federated learning under data heterogeneity}

\author{\name Bo Li\thanks{work done at CISPA} \email blia@dtu.dk \\
      \addr Technical University of Denmark
      \AND
      \name Yasin Esfandiari \email yasin.esfandiari@cispa.de \\
      \addr CISPA Helmholtz Center for Information Security
      \AND
      \name Mikkel N. Schmidt \email mnsc@dtu.dk\\
      \addr Technical University of Denmark 
      \AND
      \name Tommy S. Alstrøm \email tsal@dtu.dk\\
      \addr Technical University of Denmark 
      \AND
      \name Sebastian U. Stich \email stich@cispa.de \\
      \addr CISPA Helmholtz Center for Information Security
      }

\def\month{03}  
\def\year{2024} 
\def\openreview{\url{https://openreview.net/forum?id=c5o4HUypqm}} 

\begin{document}


\maketitle
\begin{abstract}
In federated learning, data heterogeneity is a critical challenge. A straightforward solution is to shuffle the clients' data to homogenize the distribution. However, this may violate data access rights, and how and when shuffling can accelerate the convergence of a federated optimization algorithm is not theoretically well understood. In this paper, we establish a precise and quantifiable correspondence between data heterogeneity and parameters in the convergence rate when a fraction of data is shuffled across clients. We discuss that shuffling can in some cases quadratically reduce the gradient dissimilarity with respect to the shuffling percentage, accelerating convergence. Inspired by the theory, we propose a practical approach that addresses the data access rights issue by shuffling locally generated synthetic data. The experimental results show that shuffling synthetic data improves the performance of multiple existing federated learning algorithms by a large margin. 
\end{abstract}

\section{Introduction}
Federated learning (FL) is emerging as a fundamental distributed learning paradigm that allows a central server model to be learned collaboratively from distributed clients without ever requesting the client data, thereby dealing with the issue of data access rights~\citep{DBLP:journals/corr/KonecnyMRR16, DBLP:journals/corr/abs-1912-04977, DBLP:journals/corr/abs-2107-06917, Sheller2020}. One of the most popular algorithms in FL is FedAvg~\citep{DBLP:journals/corr/KonecnyMRR16}:\ A server distributes a model to the participating clients, who then update it with their local data for multiple steps before communicating it to the server, where the received client models are aggregated, finishing one round of communication. Two main challenges in FedAvg optimization are 1) large data heterogeneity across clients and 2) limited available data on each client~\citep{DBLP:journals/corr/KonecnyMRR16}.

Data heterogeneity refers to differences in data distributions among participating clients~\citep{DBLP:journals/corr/abs-1912-04977}. Existing works typically characterize its impact on the convergence of FedAvg using bounds on the gradient dissimilarity, which suggests a slow convergence when the data heterogeneity is high~\citep{DBLP:journals/corr/abs-1910-06378, DBLP:journals/corr/abs-2003-10422}. Many efforts have been made to improve FedAvg's performance in this setting using advanced techniques such as control variates~\citep{DBLP:journals/corr/abs-1910-06378, DBLP:journals/corr/abs-2111-04263, DBLP:journals/corr/abs-2212-02191} and regularizers~\citep{DBLP:journals/corr/abs-2007-07481}. Although these algorithms have demonstrated great success in many applications, they are insufficient under high data heterogeneity~\citep{https://doi.org/10.48550/arxiv.2207.06343}. On the other hand, 
many researchers~\citep{zhao2018federated,NEURIPS2020_45713f6f} have observed that simply augmenting the dataset on each client with a small portion of shuffled data collected from all clients can significantly accelerate the convergence of FedAvg. However, it has yet to be well understood when and, in particular, by how much shuffling can accelerate the convergence. 

Another challenge arises when the amount of available data on each client is limited, especially when the number of classes or tasks is large. High-capacity deep neural networks (DNN) usually require large-scale labelled datasets to avoid overfitting~\citep{DBLP:journals/corr/HeZRS15, DBLP:journals/corr/ShrivastavaPTSW16}. However, labelled data acquisition can be expensive and time-consuming~\citep{DBLP:journals/corr/ShrivastavaPTSW16}. Therefore, learning with conditionally generated synthetic samples has been an active area of research~\citep{sehwag2022robust, DBLP:journals/corr/abs-2008-04489, DBLP:journals/corr/ShrivastavaPTSW16, zhang2022dense, DBLP:journals/corr/abs-2105-10056, li2022federated}. Adapting the framework of using synthetic data from centralized learning to FL is not trivial due to the unique properties of FL, such as distributed datasets, high communication costs, and privacy requirements~\citep{DBLP:journals/corr/abs-1912-04977}. 

In this work, we rigorously study the correspondence between the data heterogeneity and the parameters in the convergence rate via shuffling. Specifically, we show that reducing data heterogeneity by shuffling in a small percentage of data from existing clients can give a quadratic reduction in the gradient dissimilarity in some scenarios and can lead to a super-linear reduction (a factor greater than the shuffle percentage) in the number of rounds to reach the target accuracy. Further, we show that it holds in both strongly-convex functions and non-convex DNN-based FL.

\begin{figure}[tb!]
    \centering
\includegraphics[width=1.0\textwidth]{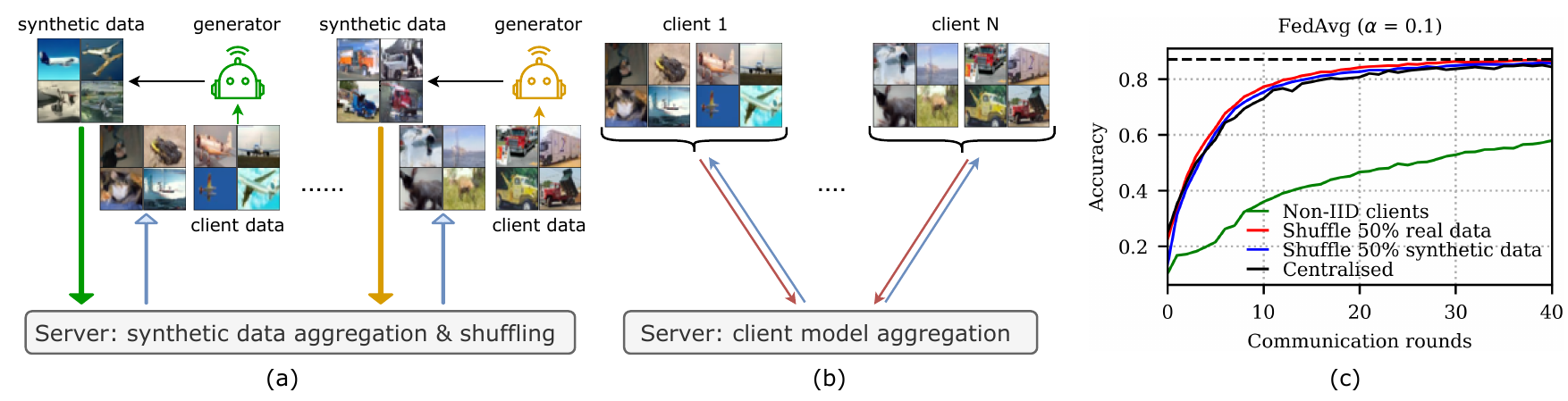}
    \caption{Our proposed framework. (a) Each client learns a generator with a subset of its local data and generates synthetic data, which are communicated to the server. The server then shuffles and sends the partitioned collection of the synthetic data to each client. (b) With the updated local data, any FL algorithms can be used to learn a server model. (c) When the clients are very heterogeneous, compared to shuffling the real data, shuffling synthetic data achieves a similar accuracy while alleviating information leakage.}
    \label{fig:main_framework}
\end{figure}

While we theoretically understand the optimization improvement from shuffling, gathering real data from clients goes against the goal of protecting privacy in FL. Therefore, we propose a more practical framework $\texttt{Fedssyn}$ where we shuffle a collection of synthetic data from all the clients at the beginning of FL (see Fig.~\ref{fig:main_framework}). Each client learns a client-specific generator with a subset of its local data and generates synthetic data. The server then receives, shuffles, partitions, and redistributes the synthetic data to each client. Collecting visually interpretable synthetic data can be more transparent and secure than collecting the black-box generators. This transforms the issue into one concerning the assurance of privacy in the generative model, an area of active research that already boasts some well-established techniques~\cite{yoon2018pategan, DBLP:conf/nips/ChenOF20, wang2023fed}. Therefore, we also illustrate the benefit of \texttt{Fedssyn} in a differentially private setting with a small-scale experiment. \looseness=-1


\textbf{Contributions:} We summarize our main results below:
\begin{itemize}[leftmargin=12pt]
    \item We rigorously decompose the impact of shuffling on the convergence rate in FedAvg. Our careful discussion provides insights in understanding when and how shuffling can accelerate the convergence of FedAvg. We empirically verify our theoretical statements on strongly convex and DNN-based non-convex functions.
    \item Inspired by the theoretical understanding of the effects of data shuffling, we present a practical framework Fedssyn, that shuffles for privacy reasons locally generated \emph{synthetic data} and can be coupled with \emph{any} FL algorithms.
    \item We empirically demonstrate that using \texttt{Fedssyn} on top of several popular FL algorithms reduces communication cost and improves the Top-1 accuracy. Results hold across multiple datasets, different levels of data heterogeneity, number of clients, and participation rates.
\end{itemize}

\subsection{Related work}

\textbf{Federated optimization:} FL is a fast-growing field~\citep{ DBLP:journals/corr/abs-2107-06917}. We here mainly focus on works that address the optimization problems in FL. FedAvg~\cite{DBLP:journals/corr/KonecnyMRR16} is one of the most commonly used optimization techniques. Despite its success in many applications, its convergence performance under heterogeneous data is still an active area of research~\citep{DBLP:journals/corr/abs-1910-06378, wang2022unreasonable, DBLP:journals/corr/abs-1909-04746, DBLP:journals/corr/abs-1812-06127, NEURIPS2020_45713f6f, DBLP:journals/corr/abs-2212-02191}. Theoretical results suggest that the \textit{drift} caused by the data heterogeneity has a strong negative impact on FedAvg convergence. 
Several lines of work have been proposed to reduce the impact of data heterogeneity, including regularizers~\citep{DBLP:journals/corr/abs-1812-06127}, control variates for correcting gradients~\citep{DBLP:journals/corr/abs-1910-06378, DBLP:journals/corr/abs-2111-04263, DBLP:journals/corr/abs-2212-02191, NEURIPS2021_7a6bda9a}, and decoupled representation and classification~\citep{https://doi.org/10.48550/arxiv.2205.13692, DBLP:journals/corr/abs-2106-05001, https://doi.org/10.48550/arxiv.2207.06343}, sophisticated aggregation strategies~\citep{DBLP:conf/nips/LinKSJ20, DBLP:journals/corr/abs-2008-03606, DBLP:journals/corr/abs-2007-07481, DBLP:journals/corr/abs-2003-00295}. However, such methods fall short in addressing high client heterogeneity and fail to achieve accuracies on par with those in the IID setting~\citep{https://doi.org/10.48550/arxiv.2207.06343}

\textbf{Synthetic data-based FL:} Many works attempt to improve the performance of FL algorithms using synthetic data~\citep{ DBLP:journals/corr/abs-2105-10056,DBLP:journals/corr/abs-2008-04489}. We can group them into 1) client-generator $\mathcal{G}_i$~\citep{xiong2023feddm, li2022federated,wang2023fed}, where a generator is trained locally and sent to the server for synthetic data generation, and 2) server-generator $\mathcal{G}$~\citep{DBLP:journals/corr/abs-2105-10056,zhang2022dense}, where a centralized generator helps with updating the server and/or client model. See Table below for a summary. Transmitting the synthetic data at every round $\tilde{\mathcal{D}}(r)$ can incur extra communication costs. Updating the server model with the synthetic data $\xx\leftarrow\tilde{\mathcal{D}}$ requires the synthetic data to be of high quality~\cite{DBLP:conf/nips/LinKSJ20, sehwag2022robust}, as low-quality synthetic data or synthetic data with a different distribution than the actual local data may discard the valuable information the server model has learned. Sending generators instead of synthetic data may be less secure, seeing that having access to the generator's parameters gives the server more powerful attacking abilities~\cite{WANG202214}.
\vspace*{-0.3cm}
\begin{table}[H]
\centering
\begin{tabular}{@{}lccccc@{}}
\toprule
       & Local $\mathcal{G}_i$ & Server $\mathcal{G}$ & $\tilde{\mathcal{D}}(r)$ & $\xx \leftarrow \tilde{\mathcal{D}}$ & Pseudo-label \\ \midrule
FedDM~\cite{xiong2023feddm}  &   $\checkmark$   & \textcolor{red}{\xmark}   &   $\checkmark$   & $\checkmark$  & \textcolor{red}{\xmark}   \\
FedGEN~\cite{DBLP:journals/corr/abs-2105-10056} & \textcolor{red}{\xmark}            &  $\checkmark$  &  $\checkmark$ & \textcolor{red}{\xmark} & \textcolor{red}{\xmark} \\
Dense~\cite{zhang2022dense}  & \textcolor{red}{\xmark}            &   $\checkmark$  &\textcolor{red}{\xmark}    &    $\checkmark$ & \textcolor{red}{\xmark}     \\
SDA-FL~\cite{li2022federated} & $\checkmark$  &  \textcolor{red}{\xmark}          & $\checkmark$ & $\checkmark$ &$\checkmark$\\ \hline
ours   &  $\checkmark$  & \textcolor{red}{\xmark}           &  \textcolor{red}{\xmark}                                             &     \textcolor{red}{\xmark}                                       & \textcolor{red}{\xmark}             \\ \bottomrule
\end{tabular}
\end{table}
\vspace*{-0.2cm}
\textbf{Generative models:} Generative models can be trained to mimic the underlying distribution of complex data modalities~\citep{NIPS2014_5ca3e9b1}. Generative Adversary Network (GANs) have achieved great success in generating high-quality images~\citep{sinha2021negative, DBLP:journals/corr/abs-2006-10738, DBLP:journals/corr/abs-2006-06676}; however, training GANs in a FL setup is challenging as each client only has limited data~\citep{NIPS2016_8a3363ab, DBLP:journals/corr/KonecnyMRR16}. Diffusion models~\citep{ho2020denoising} have attracted much attention due to the high diversity of the generated samples~\cite{sehwag2022robust} and training efficiency~\citep{ho2020denoising}. Therefore, we study to use DDPM as our local generator in this paper.

\textbf{Differentially private learning in neural networks:} Differentially private stochastic gradient descent (DP-SGD) bounds the influence of any single input on the output of machine learning algorithms to preserve privacy~\citep{brendan2018learning, 10.1145/2810103.2813687, abadi2016deep, DBLP:journals/corr/abs-1802-06739, DBLP:conf/nips/ChenOF20, yoon2018pategan}. For example,~\cite{dockhorn2022differentially} have proposed to train diffusion models with differential privacy guarantees that can alleviate privacy leakage. 


\section{Influence of the data heterogeneity on the convergence rate}
\label{sec:theory}
We formalize the problem as minimizing the expectation of a sum of stochastic functions $F$ (e.g.\ the loss associated with each datapoint):
\[f^\star := \min_{\xx\in\mathbb{R}^d} \left[\frac{1}{N}\sum_{i=1}^N f(\xx,\mathcal{D}_i) \right], \quad f(\xx,\mathcal{D}_i) := \mathbb{E}_{\xi\sim\mathcal{D}_i}F(\xx, \xi) \,, \] 
where $\mathcal{D}_i$ represents the distribution of $\xi$ on client $i$ and we use ${\mathcal{D}} := \bigcup_i \mathcal{D}_i$ to denote the uniform distribution over the joint data.  We assume gradient dissimilarity, bounded noise, and smoothness of the function, following~\cite{DBLP:journals/corr/abs-2003-10422, DBLP:journals/corr/abs-1907-04232}:

\begin{assumption}[gradient dissimilarity]\label{assum:original_gradient_dissimilarity}We assume that there exists $\zeta^2\geq 0$ such that $\forall \xx\in \mathbb{R}^d$:
\begin{align*}\textstyle \frac{1}{N}\sum_{i=1}^N||\nabla f(\xx, \mathcal{D}_i) - \nabla f(\xx,\mathcal{D})||^2 \leq \zeta^2 \,.
\end{align*}
\end{assumption}
\begin{assumption}[stochastic noise]
\label{assump:stochastic_noise} We assume that there exist 
$\sigma_{\text{avg}}^2 \geq 0$ and $\sigma^2\geq 0$ 
such that $\forall \xx\in \mathbb{R}^d$, $i \in [N]$:
\[\mathbb{E}_{\xi \sim \mathcal{D}_i}||\nabla F(\xx,\xi) - \nabla f(\xx,\mathcal{D}_i)||^2 \leq \sigma^2\,, \quad \mathbb{E}_{\xi\sim\mathcal{D}}||\nabla F(\xx,\xi) - \nabla f(\xx,\mathcal{D})||^2 \leq \sigma_{\text{avg}}^2 \,.\] 
\end{assumption}
\begin{assumption}[smoothness]\label{assum:smooth}
We assume that each $F(\xx,\xi)$ for $\xi \in \mathcal{D}_i$, $i \in [N]$ is $L$-smooth, i.e.,:
\[||\nabla f(\xx,\mathcal{D}_i) - \nabla f(\yy,\mathcal{D}_i)|| \leq L||\xx - \yy||\,, \qquad \forall \xx, \yy\in\mathbb{R}^d\,.\]
\end{assumption}
These assumptions are standard in the literature~\cite[cf.][]{DBLP:journals/corr/abs-2003-10422,NEURIPS2020_45713f6f,khaled2020tighter}. 
We further denote by $L_{\rm avg}\leq L$ the smoothness of $f(\xx,\mathcal{D})$,
and by  $L_{\rm max}\leq L$ a uniform upper bound on the smoothness of each single $f(\xx,\mathcal{D}_i)$, for all $i \in [N]$.


\subsection{Motivation}
Data heterogeneity is unavoidable in real-world FL applications since each client often collects data individually~\citep{ DBLP:journals/corr/KonecnyMRR16}. 
One simple but effective strategy to address data heterogeneity is to replace a small fraction (e.g.\ 10\%) of the client's data with shuffled data collected from other clients. \citet{NEURIPS2020_45713f6f, zhao2018federated} have observed that shuffling can vastly reduce the reached error.
In the following, we focus on explaining this empirically tremendous optimization improvement from a theoretical perspective. However, we emphasize that this is not a practical approach, as data sharing may be diametrically at odds with the fundamentals of FL.


\subsection{Modelling Data Shuffling and Theoretical Analysis}
We formally describe the data heterogeneity scenario as follows: We assume each client has access to data with distribution $\mathcal{D}_i$ that is non-iid across clients and we denote by 
${\mathcal{D}} := \bigcup_i \mathcal{D}_i$ 
the uniform distribution over the joint data. We can now model our shuffling scenario where the server mixes and redistributes a small $p$ fraction of the data by introducing new client distributions $(1-p)\mathcal{D}_i + p \mathcal{D}$ for a shuffling parameter $p\in[0, 1]$. If $p=0$, the local data remains unchanged, and if $p=1$, the data is uniform across clients. 

We argued that in practice it might be infeasible to assume access to the distribution $\mathcal{D}$. Instead, only an approximation (e.g.\ synthetic data) might be available, which we denote by $\tilde{\mathcal{D}}$. Accordingly, we define $\tilde{\sigma}_{\text{avg}}$ similarly as Assumption~\ref{assump:stochastic_noise} but with $\xi\sim\tilde{\mathcal{D}}$ instead of $\xi\sim\mathcal{D}$ and define $\tilde{L}_{\text{avg}}$ similarly as $L_{\text{avg}}$ but on $\tilde{\mathcal{D}}$. We denote the client distribution after shuffling by $\hat{\mathcal{D}}_i := (1-p)\mathcal{D}_i + p \tilde{\mathcal{D}}$ and we quantify  the data distribution differences by a new parameter $\delta$. Note that in an idealized scenario when $\tilde{\mathcal{D}}=\mathcal{D}$, then $\delta = 0$. See Appendix~\ref{appendix_sec:lemma} for more details.

\begin{assumption}[distribution shift]\label{assum:delta}
We assume that there exists $\delta^2\geq 0$ such that $\forall \xx\in \mathbb{R}^d$:
\[
 ||\nabla f(\xx,\mathcal{D}) - \nabla f(\xx,\tilde{\mathcal{D}})||^2  \leq \delta^2 \,.
\]%
\end{assumption}
\vspace{-3mm}

We now characterize how shuffling a $p$ fraction of data to each client impacts the problem difficulty.
For clarity, we will use the index $p$ to denote how 
 stochastic noise $\hat{\sigma}_p^2$, gradient dissimilarity $\hat{\zeta}_p^2$
and the smoothness constant $L_p$ 
can be estimated for the new distributions $\hat{\mathcal{D}}_i$ resulting after shuffling.

\begin{lemma}\label{lemma:updated_stochastic_noise_and_gradient_disimilarity}
If Assumption~\ref{assum:original_gradient_dissimilarity} --~\ref{assum:delta} hold, 
then in expectation over potential randomness in selecting $\tilde{\mathcal{D}}$:
\begin{align*}
    &\!\!\mathbb{E} \hat{\sigma}_p^2 \leq (1-p)\sigma^2+p\tilde{\sigma}^2_{\text{avg}} + p(1-p)\zeta^2+p(1-p)\delta^2\,,& %
    &\mathbb{E} \hat{\zeta}_p^2 \leq (1-p)^2\zeta^2\,, & \\%
    &\!\!\mathbb{E} L_p \leq (1-p) L_{\text{max}} + p \tilde{L}_{\text{avg}}\,.
\end{align*}%
\end{lemma}

Lemma~\ref{lemma:updated_stochastic_noise_and_gradient_disimilarity} shows the effective stochastic noise, gradient dissimilarity, and function smoothness when each client $i$ contains a fraction $p$ of shuffled data (see Appendix~\ref{appendix_sec:lemma} for the proof). We observe that the gradient dissimilarity decays quadratically with respect to $p$. When $p=1$, the gradient dissimilarity is $0$ as expected, since data is iid across workers. We also characterize the impact of the distribution shift from $\mathcal{D}$ to $\tilde{\mathcal{D}}$ on the effective stochastic noise. 

We now demonstrate the impact of these parameters on the convergence rate of FedAvg. We assume that there are $N$ clients. In each communication round, each client updates the received server model for $\tau$ local steps using stochastic gradient descent (SGD). The updated model is then sent back to the server for aggregation, finishing one round of communication. We repeat this process for $R$ rounds (See Algorithm II for more information). We state the convergence rate using the convergence bounds that are well-studied and established in~\cite{DBLP:journals/corr/abs-2003-10422,NEURIPS2020_45713f6f,khaled2020tighter} and combining them with our Lemma~\ref{lemma:updated_stochastic_noise_and_gradient_disimilarity}:
\begin{corollaryp}{I}[Convergence rate after shuffling]
We consider: $p$ as the fraction of shuffled data on each client; $\tau$ as local steps; under Assumption~\ref{assum:original_gradient_dissimilarity}--~\ref{assum:delta}, for any target accuracy $\epsilon > 0$, there exists a (constant) stepsize such that the accuracy can be reached after at most $T := R\cdot\tau$ iterations (in expectation):%
\begin{align*}
\textbf{$\mu$-strongly-convex}&: 
    T=\mathcal{O}\left(\frac{\hat{\sigma}_p^2}{\mu N\epsilon} + \frac{\sqrt{L}\bigl(\tau\hat{\zeta}_p + \sqrt{\tau}\hat{\sigma}_p\bigr)}{\mu\sqrt{\epsilon}} + \frac{L\tau}{\mu}\log\frac{1}{\epsilon} \right) \,, \\
\textbf{Non-convex}&:%
T=\mathcal{O}\left(\frac{L\hat{\sigma}_p^2}{N\epsilon^2} + \frac{L\bigl(\tau\hat{\zeta}_p+\sqrt{\tau}\hat{\sigma}_p\bigr)}{\epsilon^{3/2}}+\frac{L\tau}{\epsilon} \right) \,.
\end{align*}%
\label{corrolary:rate}%
\end{corollaryp}%
\vspace*{-4mm}%

    
From Corollary~\ref{corrolary:rate}, we see that if $\hat{\sigma}_p^2 > 0$, the convergence rate is asymptotically dominated by the first term, i.e.\
$\mathcal{O}\left(\frac{\hat{\sigma}_p^2}{\mu N\epsilon}\right)$,  which can be bounded by
$\mathcal{O}\left(\frac{(1-p)\sigma^2+p\tilde{\sigma}^2_{\text{avg}}+p(1-p)\zeta^2+p(1-p)\delta^2}{N\mu\epsilon} \right)$. This shows a mixed effect between the stochastic noise, gradient dissimilarity, and the distribution shift. Shuffling can have less of an impact on the convergence in the high noise regime. 
When $\sigma^2=0$, or in general in the early phases of the training (when the target accuracy $\epsilon$ is not too small),  
for problems where Lemma~\ref{lemma:updated_stochastic_noise_and_gradient_disimilarity} holds, then the number of iterations $T$ to achieve accuracy $\epsilon$ can be super-linearly reduced by a factor larger than $p$ with the increasing ratio of the shuffled data $p$, as $T$ is proportional to $\hat{\zeta}_p$ and $\sqrt{L}_p$ in expectation for strongly convex functions. We next present a numerical example to verify this.

\subsection{Illustrative experiments on convex functions}

\begin{wrapfigure}{r}{0.5\textwidth}
\centering
\includegraphics{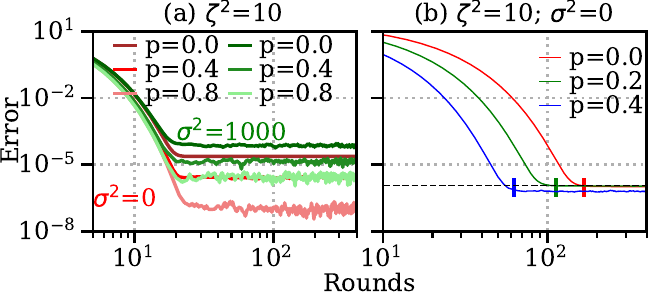}
\caption{Convergence of $\frac{1}{n}\sum_{i=1}^n||\xx_i^t-\xx^\star||^2$. (a) With a fixed $\zeta^2$ and step size, shuffling reduces the optimal error more when the stochastic noise is low (b) When gradient dissimilarity $\zeta^2$ dominates the convergence, we obtain a super-linear speedup in the number of rounds to reach $\epsilon$ by shuffling more data. The \texttt{vertical bar} shows the theoretical number of rounds to reach $\epsilon$. The stepsize is tuned in (b).}. 
\label{fig:synthetic_exp_short}
\vspace*{-10mm}
\end{wrapfigure}

We here formulate a quadratic example to verify that our theoretical results when we shuffle a $p$ faction of the local data. Following\footnote{In contrast to~\cite{DBLP:journals/corr/abs-2003-10422} that consider the online setting, we generate here a finite set of samples on each client.}~\cite{DBLP:journals/corr/abs-2003-10422}, we consider a distributed least squares objective $f(\xx) := \frac{1}{n}\sum_{i=1}^n\bigl[f_i(\xx):=\frac{1}{2n_i}\sum_{j=1}^{n_i}||\boldsymbol{A}_{ij}\xx - \boldsymbol{b}_{ij}||^2\bigr]$ with $\boldsymbol A_{ij}=i\boldsymbol I_d$, $\boldsymbol{\mu}_{i}\sim \mathcal{N}(0, \zeta^2 (id)^{-2}\boldsymbol{I}_d)$, and $\boldsymbol{b}_{ij}\sim \mathcal{N}(\boldsymbol{\mu}_i, \sigma^2 (id)^{-2}\boldsymbol{I}_d)$, where $\zeta^2$ controls the function similarity and $\sigma^2$ controls the stochastic noise (matching parameters in Corollary~\ref{corrolary:rate}). We depict the influence of shuffling on the convergence in Fig.~\ref{fig:synthetic_exp_short}. 

In Fig.~\ref{fig:synthetic_exp_short}(a), we observe that in the high noise regime ($\sigma^2=1000$), shuffling gives a smaller reduction on the optimal error than when $\sigma^2=0$. In Fig.~\ref{fig:synthetic_exp_short} (b), we tune the stepsize to reach to target accuracy $\epsilon=1.1\cdot10^{-6}$ with fewest rounds. Fig.~\ref{fig:synthetic_exp_short} (b) shows that the empirical speedup matches the theoretical speedup as the observed and theoretical number of rounds (\texttt{vertical bars}) to reach $\epsilon$ are very close---however, only when we replace $L$ in the theoretical bounds by $L_p$. 
This experiment shows that the practical performance of FedAvg can depend on $L_p$ (note in this experiment $\tilde{\mathcal{D}}=\mathcal{D}$), which hints that the pessimistic worst-case bounds from other works we have applied in Cor.~\ref{corrolary:rate} can potentially be improved.\looseness=-1



\section{Synthetic data shuffling}
We have theoretically and empirically verified the benefit of adding shuffled data to each client. However, collecting data from clients goes against the goal of protecting user rights in FL. To still enjoy the benefit of shuffling and alleviate information leakage, we propose a practical framework that shuffles aggregated and more interpretable synthetic images from all the clients (see Fig.~\ref{fig:main_framework}), shifting the problem into one of assuring privacy of the generator.

\vspace*{-0.4cm}
\begin{minipage}{0.51\textwidth}
\begin{algorithm}[H]
\floatname{algorithm}{Algorithm I}
\small
\caption{Fedssyn}
\label{Algorithm:fedss_algo}
\begin{algorithmic}[1]
\Procedure{Synthetic data generation}{}
    \For{ \textbf{client $i=1,\dots, N$ in parallel}}
    \State sample $\rho\cdot n_i$ data points from $\mathcal{D}_i$ 
    \State train a \textit{generator} $\mathcal{G}_i$
    \State generate $\tilde{\mathcal{D}}_i$ with $\tilde{n}$ samples
    \State Communicate $\tilde{\mathcal{D}}_i$ to the server
    \EndFor
    \State Server \textit{shuffles} the \textit{aggregated} synthetic data $\{\tilde{\mathcal{D}}_1,\dots, \tilde{\mathcal{D}}_N\}$ and \textit{split} it to $N$ parts ($\{\tilde{\mathcal{D}}_{si}\}$)
    \State Server \textit{distributes} $\tilde{\mathcal{D}}_{si}$ ($|\tilde{\mathcal{D}}_{si}|=\tilde{n}$) to each client
    \State Run FL algorithm with the updated client dataset on each client $\hat{\mathcal{D}}_i := \mathcal{D}_i\cup\tilde{\mathcal{D}}_{si}$
\EndProcedure
\Statex
\end{algorithmic}
\vspace{-0.35cm}%
\end{algorithm}
\end{minipage}
\hfill
\begin{minipage}{.46\textwidth}
\begin{algorithm}[H]
\small
\floatname{algorithm}{Algorithm II}
\caption{Federated Averaging (FedAvg)}
\label{Algorithm:fed_avg}
\begin{algorithmic}[1]
\Procedure{FedAvg}{}
    \For {$r= 1,\dots,R$}
        \State Sample clients $S \subseteq \{1,\dots,N\}$
        \State Send server model $\xx$ to all clients $i\in S$
        \For { \textbf{client $i \in S$ in parallel}}
        \State initialise local model $\yy_i \leftarrow \xx$
        \For { $k=1,\dots,\tau$}
            \State $\yy_i \leftarrow \yy_i - \eta\nabla F_i(\yy_i)$
        \EndFor 
        \EndFor
        \State $\xx \leftarrow \xx + \frac{1}{|S|}\sum_{i\in S}(\yy_i - \xx)$
    \EndFor
\EndProcedure
\Statex
\end{algorithmic}
  \vspace{-0.3cm}%
\end{algorithm}
\end{minipage}

We denote the client dataset, client generated synthetic dataset, and shuffled synthetic dataset by $\mathcal{D}_i$, $\tilde{\mathcal{D}}_i$, and $\tilde{\mathcal{D}}$ (in accordance with the notation for the data distributions in Sec.~\ref{sec:theory}). On each client $i$, we uniformly subsample $\rho\cdot n_i$ data points from the local dataset $\mathcal{D}_i$, given $\rho \in (0, 1]$ and $n_i = |\mathcal{D}_i|$, which we use to locally train a class-conditional generator. Using a subset rather than the entire dataset can alleviate the exposure of the class distribution from each client. Given the trained local generator, we generate $\tilde{\mathcal{D}}_i$ with $|\tilde{\mathcal{D}}_i|=\tilde{n}$ synthetic data, matching the class frequency in the subset. We assume that all the clients are willing to send their synthetic dataset to the server. The server then shuffles the aggregated synthetic datasets $\cup_{i\in[N]}\tilde{\mathcal{D}}_i$ and distributes the data uniformly among the clients. In practice, when the number of synthetic data is large, their statistical dependency is negligible, and the synthetic data can be considered nearly iid. Reflecting this, we denote the \textit{partitioned} shuffled synthetic dataset as $\tilde{\mathcal{D}}_{si}$ on each client, and the proportion of the synthetic dataset is calculated as $p:=\frac{\tilde{n}}{n_i + \tilde{n}}$ (see Algorithm \textbf{I}). 

We then run FL algorithms with the updated dataset on each client $\mathcal{D}_i \cup \tilde{\mathcal{D}}_{si}$. Our framework has no requirements for the FL algorithm, so we here briefly describe one of the most common FL algorithms: FedAvg~\citep{DBLP:journals/corr/KonecnyMRR16}, as shown in Algorithm \textbf{II}. FedAvg mainly has two steps: local model updating and server aggregating. We initialize the server with a model $\xx$. In each communication round, each participating client $S\subseteq[N]$ receives a copy of the server parameter $\xx$ and performs $K$ steps local updates (e.g., $K$ steps SGD). The updated model $\yy_i$ is then sent to the server. FedAvg calculates the server model by averaging over the received client models, finishing one round of communication. We repeat this for $R$ rounds or until the target accuracy is achieved.

\textbf{Advantages of sending synthetic data rather than a generator for one-round:} 1) sending synthetic data can be more secure than sending the generator to the server since having access to generator's parameters gives the server more powerful attacking abilities~\citep{WANG202214} 2) the clients can easily inspect the information shared with the server as the synthetic images are visually more interpretable than black-box generators 3) one-round synthetic data communication can be more secure than multi-round generator/synthetic data transmission~\citep{xiong2023feddm,DBLP:journals/corr/abs-2105-10056} as the server cannot infer the local data distribution by identifying the difference between the updated synthetic data 4) depending on the generator, sending synthetic data can incur less communication cost, for example, transmitting the synthetic data leads to a $2\times$-$30\times$ reduction on the number of parameters communicated than transmitting the generator in our experiment.

\section{Experimental setup}
We show the effectiveness of our proposed method on CIFAR10 and CIFAR100~\citep{Krizhevsky2009LearningML} image classification tasks. 
We partition the training dataset using Dirichlet distribution with a concentration parameter $\alpha$ to simulate the heterogeneous scenarios following~\cite{DBLP:conf/nips/LinKSJ20}. Smaller $\alpha$ corresponds to higher data heterogeneity. We pick $\alpha \in \{0.01, 0.1\}$ as they are commonly used~\citep{https://doi.org/10.48550/arxiv.2207.06343, DBLP:conf/nips/LinKSJ20}. Each client has a local dataset, kept fixed and local throughout the communication rounds. Experiments using MNIST and dSprites are in Appendix~\ref{appendix_sec:mnist} and~\ref{appendix_sec:dsprite}.

We use class-conditional DDPM~\cite{ho2020denoising} on each client. We assume that all the clients participate in training DDPM by using 75\% of their local data as the training data. Each client trains DDPM with a learning rate of $0.0001$, $1000$ diffusion time steps, $256$ batch size, and $500$ epochs. These hyperparameters are the same for all experiments. Once the training finishes, each client simulates $\tilde{n}=\frac{50000}{N}$ (there are 50K training images in CIFAR10 and CIFAR100) synthetic images with the label, which are then sent to the server. The server shuffles the aggregated synthetic data and then distributes it to each client equally.

We then perform federated optimization. We experiment with some of the popular FL algorithms including FedAvg~\citep{DBLP:journals/corr/KonecnyMRR16}, FedProx~\citep{DBLP:journals/corr/abs-1812-06127}, SCAFFOLD~\citep{DBLP:journals/corr/abs-1910-06378}, FedDyn~\citep{DBLP:journals/corr/abs-2111-04263}, and FedPVR~\citep{DBLP:journals/corr/abs-2212-02191}. We use VGG-11 for all the experiments. We train the FL algorithm with and without using the synthetic dataset to demonstrate the speedup obtained from using synthetic dataset quantitatively. 

We use $N\in \{10, 40, 100\}$ as the number of clients and $C \in \{0.1, 0.2, 0.4, 1.0\}$ as the participation rate. For partial participation, we randomly sample $N\cdot C$ clients per communication round. We use a batch size of $256$, $10$ local epochs, and the number of gradient steps $\frac{10n_i}{256}$. We tune the learning rate from $\{0.01, 0.05, 0.1\}$ with the local validation dataset. The results are given as an average of three repeated experiments with different random initializations. Note that the reported accuracy is calculated on the real server test dataset.

\section{Experimental results}
We show the performance of \texttt{Fedssyn} here. Our main findings are: 1) Sampling shuffled synthetic data into each client can significantly reduce the number of rounds used to reach a target accuracy (1.6x-23x) and increase the Top-1 accuracy. 2) Fedssyn can reduce the number of communicated parameters to reach the target accuracy up to 95\% than vanilla FedAvg. 3) Fedssyn is more robust across different levels of data heterogeneity compared to other synthetic-data based approaches 4) The quantified gradient dissimilarity and stochastic noise in DNN match the theoretical statements

\textbf{Improved communication efficiency}
We first report the required number of communication rounds to reach the target accuracy in Table~\ref{tab:summary_result}. The length of the grey and red colour bar represents the required number of rounds to reach accuracy $m$ when the local dataset is $\mathcal{D}_i$ and $\mathcal{D}_i\cup\tilde{\mathcal{D}}_{si}$, respectively. The speedup is the ratio between the number of rounds in these two settings. We observe that adding shuffled synthetic data to each client can reduce the number of communication rounds to reach the target accuracy for all the cases. See Table~\ref{appendix_table:extension_of_the_main_table} for the actual number of rounds required to reach $m$-accuracy.. 
\vspace*{-3mm}
\begin{table}[ht!]
\setlength{\tabcolsep}{2pt}
\caption{The required number of rounds to reach target accuracy $m$. $m$ is chosen such that most of the FL algorithms can reach such accuracy within 100 rounds. The length of the grey and red bar equals to the number of rounds for experiments without the synthetic dataset (\textcolor{black!20}{$\mathcal{D}_i$}) and with the shuffled synthetic dataset (\textcolor{red}{$\mathcal{D}_i\cup\tilde{\mathcal{D}}_{si}$}), respectively. We annotate the length of the red bar, and the speedup (x) is the ratio between the number of rounds in these two settings. ``$>$'' means experiments with $\mathcal{D}_i$ as the local data cannot reach accuracy $m$ within 100 communication rounds. \textbf{Using shuffled synthetic dataset reduces the required number of communication round to reach the target accuracy in all cases.}}%
\vspace*{-3mm}
\label{tab:summary_result}
\resizebox{0.74\textwidth}{!}
{\begin{minipage}{\textwidth}
\begin{tabular}{@{}lllllllll@{}}
\toprule
              & \multicolumn{2}{c}{Full participation}              & \multicolumn{6}{c}{Partial participation} \\ 
              \cmidrule(lr){2-3}\cmidrule(l){4-9} & \multicolumn{2}{c}{N=10} &  \multicolumn{2}{c}{N=10 (C=40\%)} & \multicolumn{2}{c}{N=40 (C=20\%)} & \multicolumn{2}{c}{N=100 (C=10\%)} \\\cmidrule(lr){2-3}\cmidrule(l){4-5}\cmidrule(lr){6-7}\cmidrule(l){8-9}
 &$\alpha=0.01$ & $\alpha=0.1$ & $\alpha=0.01$  & $\alpha=0.1$   & $\alpha=0.01$           & $\alpha=0.1$ & $\alpha=0.01$           & $\alpha=0.1$           \\ \cmidrule(lr){2-9}
 & \multicolumn{7}{c}{CIFAR10} \\ \cmidrule(lr){2-9}
 & m = 44\% & m = 66\% & m = 44\% & m = 66\% & m = 44\% & m = 66\% & m = 44\% & m = 66\%\\
FedAvg & \progressbar[linecolor=white, filledcolor=red, ticksheight=0.0, heighta=8pt, width=2.225em]{0.0449}{\raisebox{1.5pt}{$4(22\text{x})$}}            &  \progressbar[linecolor=white, filledcolor=red, ticksheight=0.0, heighta=8pt, width=1.7em]{0.12}{\raisebox{1.5pt}{$8(8.5\text{x})$}}  &   \progressbar[linecolor=white, filledcolor=red, ticksheight=0.0, heighta=8pt, width=2.5em]{0.04}{\raisebox{1.5pt}{$4(>25\text{x})$}} &  \progressbar[linecolor=white, filledcolor=red, ticksheight=0.0, heighta=8pt, width=1.125em]{0.177}{\raisebox{1.5pt}{$8(5.6\text{x})$}} &  \progressbar[linecolor=white, filledcolor=red, ticksheight=0.0, heighta=8pt, width=2.5em]{0.12}{\raisebox{1.5pt}{$12(8.3\text{x})$}}    & \progressbar[linecolor=white, filledcolor=red, ticksheight=0.0, heighta=8pt, width=1.475em]{0.36}{\raisebox{1.5pt}{$21(2.8\text{x})$}}   &\progressbar[linecolor=white, filledcolor=red, ticksheight=0.0, heighta=8pt, width=2.33em]{0.20}{\raisebox{1.5pt}{$19(4.9\text{x})$}} & \progressbar[linecolor=white, filledcolor=red, ticksheight=0.0, heighta=8pt, width=2.50em]{0.38}{\raisebox{1.5pt}{$38(>2.6\text{x})$}}\\
Scaffold       &  \progressbar[linecolor=white, filledcolor=red, ticksheight=0.0, heighta=8pt, width=0.925em]{0.101}{\raisebox{1.5pt}{$4(9.3\text{x})$}}            & \progressbar[linecolor=white, filledcolor=red, ticksheight=0.0, heighta=8pt, width=1.07em]{0.16}{\raisebox{1.5pt}{$7(6.1\text{x})$}}  & \progressbar[linecolor=white, filledcolor=red, ticksheight=0.0, heighta=8pt, width=2.5em]{0.05}{\raisebox{1.5pt}{$5(>20\text{x})$}}  & \progressbar[linecolor=white, filledcolor=red, ticksheight=0.0, heighta=8pt, width=0.65em]{0.384}{\raisebox{1.5pt}{$10(2.6\text{x})$}}  & \progressbar[linecolor=white, filledcolor=red, ticksheight=0.0, heighta=8pt, width=2.05em]{0.1463}{\raisebox{1.5pt}{$12(6.8\text{x})$}} & \progressbar[linecolor=white, filledcolor=red, ticksheight=0.0, heighta=8pt, width=1.35em]{0.37}{\raisebox{1.5pt}{$20(2.7\text{x})$}}  & \progressbar[linecolor=white, filledcolor=red, ticksheight=0.0, heighta=8pt, width=2.50em]{0.19}{\raisebox{1.5pt}{$19(>5.3\text{x})$}} & \progressbar[linecolor=white, filledcolor=red, ticksheight=0.0, heighta=8pt, width=2.10em]{0.37}{\raisebox{1.5pt}{$31(2.7\text{x})$}}             \\
FedProx  & \progressbar[linecolor=white, filledcolor=red, ticksheight=0.0, heighta=8pt, width=2.35em]{0.04}{\raisebox{1.5pt}{$4(23.5\text{x})$}}            &  \progressbar[linecolor=white, filledcolor=red, ticksheight=0.0, heighta=8pt, width=1.05em]{0.19}{\raisebox{1.5pt}{$8(5.3\text{x})$}}    & \progressbar[linecolor=white, filledcolor=red, ticksheight=0.0, heighta=8pt, width=2.5em]{0.04}{\raisebox{1.5pt}{$4(>25\text{x})$}}                      & \progressbar[linecolor=white, filledcolor=red, ticksheight=0.0, heighta=8pt, width=1.85em]{0.12}{\raisebox{1.5pt}{$9(8.2\text{x})$}} & \progressbar[linecolor=white, filledcolor=red, ticksheight=0.0, heighta=8pt, width=2.5em]{0.09}{\raisebox{1.5pt}{$9(>11.1\text{x})$}}                 & \progressbar[linecolor=white, filledcolor=red, ticksheight=0.0, heighta=8pt, width=1.77em]{0.21}{\raisebox{1.5pt}{$15(4.7\text{x})$}} &\progressbar[linecolor=white, filledcolor=red, ticksheight=0.0, heighta=8pt, width=2.50em]{0.31}{\raisebox{1.5pt}{$31(>3.2\text{x})$}} & \progressbar[linecolor=white, filledcolor=red, ticksheight=0.0, heighta=8pt, width=2.50em]{0.37}{\raisebox{1.5pt}{$37(>2.7\text{x})$}}  \\
FedDyn        & \progressbar[linecolor=white, filledcolor=red, ticksheight=0.0, heighta=8pt, width=1.225em]{0.0612}{\raisebox{1.5pt}{$3(16.3\text{x})$}}              & \progressbar[linecolor=white, filledcolor=red, ticksheight=0.0, heighta=8pt, width=1.85em]{0.067}{\raisebox{1.5pt}{$5(14.8\text{x})$}}             & \progressbar[linecolor=white, filledcolor=red, ticksheight=0.0, heighta=8pt, width=1.675em]{0.059}{\raisebox{1.5pt}{$4(16.8\text{x})$}}   & \progressbar[linecolor=white, filledcolor=red, ticksheight=0.0, heighta=8pt, width=1.875em]{0.093}{\raisebox{1.5pt}{$7(10.7\text{x})$}}  &\progressbar[linecolor=white, filledcolor=red, ticksheight=0.0, heighta=8pt, width=2.5em]{0.07}{\raisebox{1.5pt}{$7(>14.3\text{x})$}}                  & \progressbar[linecolor=white, filledcolor=red, ticksheight=0.0, heighta=8pt, width=1.575em]{0.19}{\raisebox{1.5pt}{$12(>5.3\text{x})$}}  & \progressbar[linecolor=white, filledcolor=red, ticksheight=0.0, heighta=8pt, width=2.50em]{0.17}{\raisebox{1.5pt}{$17(>5.9\text{x})$}} & \progressbar[linecolor=white, filledcolor=red, ticksheight=0.0, heighta=8pt, width=2.50em]{0.31}{\raisebox{1.5pt}{$31(>3.2\text{x})$}}                   \\
FedPVR        & \progressbar[linecolor=white, filledcolor=red, ticksheight=0.0, heighta=8pt, width=1.925em]{0.05}{\raisebox{1.5pt}{$3(19.3\text{x})$}}  & \progressbar[linecolor=white, filledcolor=red, ticksheight=0.0, heighta=8pt, width=0.75em]{0.23}{\hspace{0.5em}\raisebox{1.5pt}{$7(4.3\text{x})$}}  &  \progressbar[linecolor=white, filledcolor=red, ticksheight=0.0, heighta=8pt, width=2.5em]{0.04}{\raisebox{1.5pt}{$4(>25\text{x})$}}  & \progressbar[linecolor=white, filledcolor=red, ticksheight=0.0, heighta=8pt, width=1.9em]{0.105}{\raisebox{1.5pt}{$8(9.5\text{x})$}} & \progressbar[linecolor=white, filledcolor=red, ticksheight=0.0, heighta=8pt, width=2.125em]{0.1176}{\raisebox{1.5pt}{$10(8.5\text{x})$}}   & \progressbar[linecolor=white, filledcolor=red, ticksheight=0.0, heighta=8pt, width=1.45em]{0.36}{\raisebox{1.5pt}{$21(2.8\text{x})$}} & \progressbar[linecolor=white, filledcolor=red, ticksheight=0.0, heighta=8pt, width=2.50em]{0.21}{\raisebox{1.5pt}{$21(>4.8\text{x})$}} & \progressbar[linecolor=white, filledcolor=red, ticksheight=0.0, heighta=8pt, width=2.50em]{0.35}{\raisebox{1.5pt}{$35(>2.9\text{x})$}}               \\ \cmidrule(lr){2-9}
 & \multicolumn{7}{c}{CIFAR100} \\ \cmidrule(lr){2-9} 
 & m = 30\% & m = 40\% & m = 20\% & m = 20\% &m = 30\% & m = 40\%&m = 30\% & m = 30\% \\ 
FedAvg & \progressbar[linecolor=white, filledcolor=red, ticksheight=0.0, heighta=8pt, width=2.2em]{0.1704}{\raisebox{1.5pt}{$15(5.9\text{x})$}}            & \progressbar[linecolor=white, filledcolor=red, ticksheight=0.0, heighta=8pt, width=2.5em]{0.19}{\raisebox{1.5pt}{$19(>5.2\text{x})$}}   &\progressbar[linecolor=white, filledcolor=red, ticksheight=0.0, heighta=8pt, width=2.5em]{0.17}{\raisebox{1.5pt}{$17(>5.9\text{x})$}}   &\progressbar[linecolor=white, filledcolor=red, ticksheight=0.0, heighta=8pt, width=2.5em]{0.21}{\raisebox{1.5pt}{$21(>4.8\text{x})$}} & \progressbar[linecolor=white, filledcolor=red, ticksheight=0.0, heighta=8pt, width=2.5em]{0.37}{\raisebox{1.5pt}{$37(>2.7\text{x})$}}  & \progressbar[linecolor=white, filledcolor=red, ticksheight=0.0, heighta=8pt, width=2.5em]{0.89}{\raisebox{1.5pt}{$89(>1.1\text{x})$}} & \progressbar[linecolor=white, filledcolor=red, ticksheight=0.0, heighta=8pt, width=2.50em]{0.70}{\raisebox{1.5pt}{$70(>1.4\text{x})$}}  & \progressbar[linecolor=white, filledcolor=red, ticksheight=0.0, heighta=8pt, width=2.50em]{0.59}{\raisebox{1.5pt}{$59(>1.7\text{x})$}}                 \\
Scaffold      &  \progressbar[linecolor=white, filledcolor=red, ticksheight=0.0, heighta=8pt, width=1.025em]{0.341}{\raisebox{1.5pt}{$14(2.9\text{x})$}}            & \progressbar[linecolor=white, filledcolor=red, ticksheight=0.0, heighta=8pt, width=1.375em]{0.3454}{\raisebox{1.5pt}{$19(2.9\text{x})$}}          &   \progressbar[linecolor=white, filledcolor=red, ticksheight=0.0, heighta=8pt, width=1.35em]{0.28}{\raisebox{1.5pt}{$15(3.6\text{x})$}} & \progressbar[linecolor=white, filledcolor=red, ticksheight=0.0, heighta=8pt, width=2.5em]{0.21}{\raisebox{1.5pt}{$21(4.8\text{x})$}} & \progressbar[linecolor=white, filledcolor=red, ticksheight=0.0, heighta=8pt, width=2.5em]{0.24}{\raisebox{1.5pt}{$24(>4.2\text{x})$}} & \progressbar[linecolor=white, filledcolor=red, ticksheight=0.0, heighta=8pt, width=2.5em]{0.71}{\raisebox{1.5pt}{$71(>1.4\text{x})$}}  & \progressbar[linecolor=white, filledcolor=red, ticksheight=0.0, heighta=8pt, width=2.50em]{0.61}{\raisebox{1.5pt}{$61(1.6\text{x})$}} & \progressbar[linecolor=white, filledcolor=red, ticksheight=0.0, heighta=8pt, width=2.50em]{0.52}{\raisebox{1.5pt}{$52(>1.9\text{x})$}}                  \\
FedProx  & \progressbar[linecolor=white, filledcolor=red, ticksheight=0.0, heighta=8pt, width=2.5em]{0.17}{\raisebox{1.5pt}{$17(5.9\text{x})$}}              & \progressbar[linecolor=white, filledcolor=red, ticksheight=0.0, heighta=8pt, width=2.5em]{0.23}{\raisebox{1.5pt}{$23(4.3\text{x})$}}             &    \progressbar[linecolor=white, filledcolor=red, ticksheight=0.0, heighta=8pt, width=2.5em]{0.18}{\raisebox{1.5pt}{$17(>5.6\text{x})$}}                       & \progressbar[linecolor=white, filledcolor=red, ticksheight=0.0, heighta=8pt, width=2.5em]{0.23}{\raisebox{1.5pt}{$23(>4.3\text{x})$}} & \progressbar[linecolor=white, filledcolor=red, ticksheight=0.0, heighta=8pt, width=2.5em]{0.33}{\raisebox{1.5pt}{$33(>3.0\text{x})$}} & \progressbar[linecolor=white, filledcolor=red, ticksheight=0.0, heighta=8pt, width=2.50em]{1.00}{\raisebox{1.5pt}{$100+ (-)$}} &\progressbar[linecolor=white, filledcolor=red, ticksheight=0.0, heighta=8pt, width=2.50em]{0.75}{\raisebox{1.5pt}{$75(>1.3\text{x})$}} & \progressbar[linecolor=white, filledcolor=red, ticksheight=0.0, heighta=8pt, width=2.50em]{0.60}{\raisebox{1.5pt}{$60(>1.7\text{x})$}}                  \\
FedDyn        &  \progressbar[linecolor=white, filledcolor=red, ticksheight=0.0, heighta=8pt, width=0.9em]{0.277}{\raisebox{1.5pt}{$10(3.6\text{x})$}}
 &\progressbar[linecolor=white, filledcolor=red, ticksheight=0.0, heighta=8pt, width=1.65em]{0.182}{\raisebox{1.5pt}{$12(5.5\text{x})$}}             &   \progressbar[linecolor=white, filledcolor=red, ticksheight=0.0, heighta=8pt, width=2.5em]{0.13}{\raisebox{1.5pt}{$13(>7.7\text{x})$}}                       &\progressbar[linecolor=white, filledcolor=red, ticksheight=0.0, heighta=8pt, width=2.5em]{0.15}{\raisebox{1.5pt}{$15(6.7\text{x})$}} & \progressbar[linecolor=white, filledcolor=red, ticksheight=0.0, heighta=8pt, width=2.5em]{0.25}{\raisebox{1.5pt}{$25(>4\text{x})$}} & \progressbar[linecolor=white, filledcolor=red, ticksheight=0.0, heighta=8pt, width=2.5em]{0.78}{\raisebox{1.5pt}{$78(>1.3\text{x})$}} & \progressbar[linecolor=white, filledcolor=red, ticksheight=0.0, heighta=8pt, width=2.50em]{0.61}{\raisebox{1.5pt}{$61(>1.6\text{x})$}} & \progressbar[linecolor=white, filledcolor=red, ticksheight=0.0, heighta=8pt, width=2.50em]{0.60}{\raisebox{1.5pt}{$60(>1.7\text{x})$}}                       \\
FedPVR        & \progressbar[linecolor=white, filledcolor=red, ticksheight=0.0, heighta=8pt, width=2.15em]{0.1627}{\raisebox{1.5pt}{$14(6.1\text{x})$}}  &    \progressbar[linecolor=white, filledcolor=red, ticksheight=0.0, heighta=8pt, width=0.95em]{0.5}{\raisebox{1.5pt}{$19(2.0\text{x})$}}   &     \progressbar[linecolor=white, filledcolor=red, ticksheight=0.0, heighta=8pt, width=2.5em]{0.15}{\raisebox{1.5pt}{$15(>6.6\text{x})$}} & \progressbar[linecolor=white, filledcolor=red, ticksheight=0.0, heighta=8pt, width=2.5em]{0.2}{\raisebox{1.5pt}{$20(5.0\text{x})$}}  & \progressbar[linecolor=white, filledcolor=red, ticksheight=0.0, heighta=8pt, width=2.5em]{0.23}{\raisebox{1.5pt}{$23(>4.3\text{x})$}}    & \progressbar[linecolor=white, filledcolor=red, ticksheight=0.0, heighta=8pt, width=2.5em]{0.59}{\raisebox{1.5pt}{$59(>1.7\text{x})$}} & \progressbar[linecolor=white, filledcolor=red, ticksheight=0.0, heighta=8pt, width=2.50em]{0.71}{\raisebox{1.5pt}{$71(>1.4\text{x})$}} & \progressbar[linecolor=white, filledcolor=red, ticksheight=0.0, heighta=8pt, width=2.50em]{0.58}{\raisebox{1.5pt}{$58(>1.7\text{x})$}}  \\\bottomrule 
\end{tabular}
\end{minipage}}
\end{table}

\textbf{Reduced communication cost} Following~\cite{DBLP:journals/corr/abs-2111-04263}, we define the communication cost as the total number of parameters communicated between the clients and the server to reach a target accuracy. Assume the size of the synthetic data is $M_s$ and the size of the FL model is $M_c$, then the communication cost is calculated as $2M_s+2RM_c$, where $R$ is the number of rounds to reach the target accuracy as shown in Table~\ref{tab:summary_result}. In our experiment, $M_s$ is 2x-24x smaller than $M_c$, together with the drastically reduced number of communication rounds as shown in Table~\ref{tab:summary_result}, we can save the communication cost up to 95\%. See Table~\ref{appendix_table:communication_cost} in the Appendix for the communication cost. 

\textbf{Higher Top-1 accuracy} To verify that the accuracy improvement is mainly due to the changes in the convergence parameters rather than the increased number of images on each client, we compare experiments where the local dataset is $\mathcal{D}_i$, the mixture between local real and local synthetic data $\mathcal{D}_i\cup \tilde{\mathcal{D}}_i$, the mixture between local real and shuffled synthetic data $\mathcal{D}_i\cup\tilde{\mathcal{D}}_{si}$. 
\begin{figure}[ht!]
    \centering
    \includegraphics{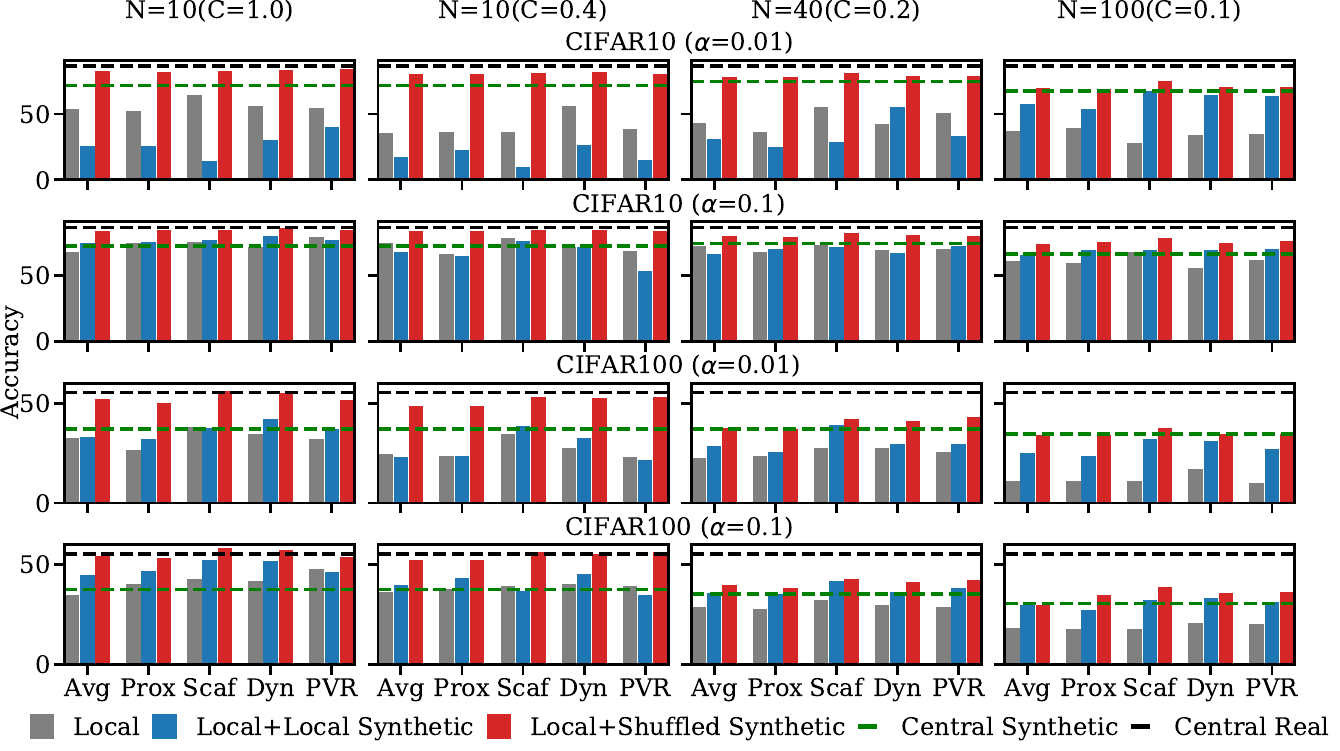}
    \caption{Top-1 accuracy. We compare the experiments where the local dataset is \textcolor{darkgray}{$\mathcal{D}_i$}, \textcolor{MidnightBlue}{$(1-p)\mathcal{D}_i+p\tilde{\mathcal{D}}_i$ (local+local synthetic data)}, and \textcolor{BrickRed}{$(1-p)\mathcal{D}_i+p\tilde{\mathcal{D}}_{si}$ (local+shuffled synthetic data)}. The black and green dotted lines represent the accuracy using the centralised real and synthetic data, respectively. Using shuffled synthetic data (\textcolor{BrickRed}{red bar}) boosts the Top-1 accuracy, and in some cases, even matches the centralised accuracy.}
    \label{fig:top_1_accuracy}
\end{figure}

In Fig.~\ref{fig:top_1_accuracy}, we observe that using the shuffled synthetic data can significantly improve the Top-1 accuracy in nearly all the cases compared to other baselines, indicates that the vast performance improvement is mainly due to the changes in the parameters in the convergence rate. We see that adding local synthetic data can sometimes lower the accuracy (e.g. \texttt{local+local synthetic} in CIFAR10 ($\alpha=0.01$)). This is primarily because the local synthetic data amplifies the heterogeneity further and makes it more challenging for an FL algorithm to converge. Additionally, the large jump from using local synthetic data when the number of clients is large (e.g. CIFAR100 with 100 clients) is mainly because the increased number of data points per client can assist in local training. We observe a smaller performance improvement when the number of clients is large ($N=100$) using Fedssyn, we can however further improve this by generating and shuffling more synthetic images.

\begin{figure}[ht!]
    \centering
    \includegraphics[width=1.0\textwidth]{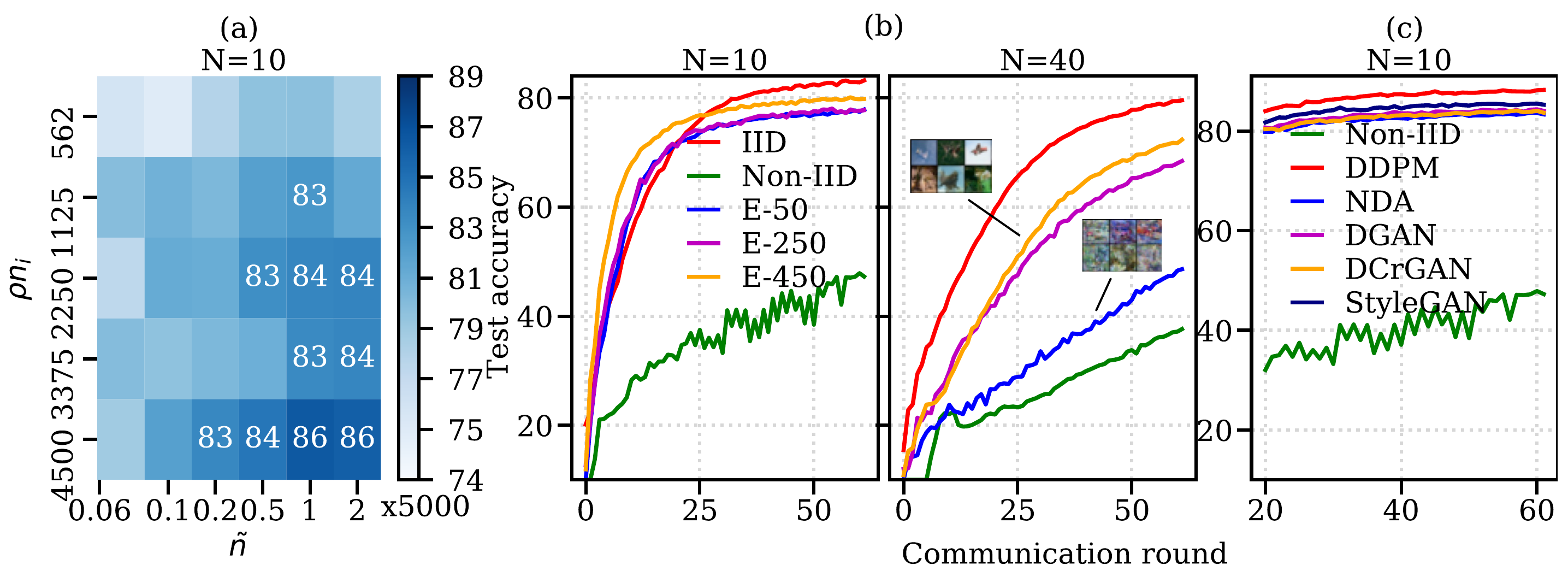}
    \vspace*{-4mm}
    \caption{Sensitivity analysis using FedAvg and CIFAR10: (a) the influence of the number of images used for training the generator $\rho\cdot n_i$ and the number of synthetic images per client $\tilde{n}$ ($\alpha=0.1$). We annotate the combination that performs better than the centralized baseline (b) influence of the number of training epochs for the generator with 10 and 40 clients ($\alpha$=0.01). (c) the influence of using different generators ($\alpha=0.01$). We obtain better performance using DDPM than other generators.}
    \label{fig:ablation_figure}
\end{figure}

\textbf{Sensitivity analysis} We here depict the sensitivity of FedAvg on the quality of the synthetic data using CIFAR10 in Fig.~\ref{fig:ablation_figure}. Fig.~\ref{fig:ablation_figure} (a) shows that the number of images for training DDPM has a substantial impact on FL performance, which is reasonable, as we can acquire higher-quality synthetic images if we use more images to train the generator. Fig.~\ref{fig:ablation_figure} (b) shows that when the number of clients is low, the synthetic images extracted from the early training checkpoints (\texttt{E=50}) already have high quality. However, when the number of clients is high, the longer we train the generator, the better FL performance we can achieve, i.e., we obtain significantly better accuracy when we train the generator for 450 epochs (\texttt{E-450}) than 50 epochs (\texttt{E-50}). Note, in our experimental setup, a higher number of clients means $n_i=|\mathcal{D}_i|$ is smaller as $n_i = \frac{50,000}{N}$ where $N$ is the number of clients. Fig.~\ref{fig:ablation_figure} (c) shows that using DDPM is slightly better than GAN-based generators in terms of accuracy.

\textbf{Comparison against other synthetic-data based approaches} We compare Fedssyn (FedAvg+shuffled synthetic data) with FedGEN~\citep{DBLP:journals/corr/abs-2105-10056} and DENSE~\citep{zhang2022dense} in Table~\ref{tab:comparison_with_other_methods}. We train FedGEN and DENSE using the default hyperparameters as described in ~\cite{DBLP:journals/corr/abs-2105-10056, zhang2022dense}. Following the setup in DENSE~\citep{zhang2022dense}, we use 10 clients with full participation, which is a commonly used FL setup in the literature~\citep{DBLP:journals/corr/abs-2212-02191, DBLP:journals/corr/abs-2103-16257, https://doi.org/10.48550/arxiv.2207.06343}. We can potentially improve the performance with more careful tuning. Comparably, Fedssyn achieves better Top-1 performance and is more robust against data heterogeneity levels. Compared to FedGEN which transmits the generator in every round, we only communicate the synthetic data once, which can be more secure and consume less cost. For example, we calculate the ratio between the required memory cost to reach the target accuracy for Fedssyn and FedGEN (last column in Table~\ref{tab:comparison_with_other_methods}) with $\frac{(2M_c + M_g)R_g}{2M_cR + 2M_s}$, where $M_g$ is the size of the generator ($\sim1.2-1.4$MB), $R_g$ and $R$ are the number of round to reach the target accuracy using FedGEN and Fedssyn, respectively, and $M_c=37.2$MB is the size of the classification model, $M_s=15.5$MB is the size of the transmitted synthetic data. We consume less communication cost than FedGEN to reach the same level of the target accuracy.See Appendix~\ref{appendix_sec:related_work_comparison} for more information regarding the training details and Fig.~\ref{fig:test_accuracy_for_fed_gen} for the accuracy over communication rounds.

\textbf{Parameters in the convergence rate} We investigate the impact of using shuffled synthetic data on the parameters in the convergence rate for DNN-based FedAvg in Fig.~\ref{fig:influence_of_sigma_zeta_main_text}. We use CIFAR10, 10 clients with full participation, $\alpha=0.1$. We observe that $(1-p)\sigma^2$ dominates over other terms in the effective stochastic noise, which means the first term in the convergence rate for non-convex function in Corollary~\ref{corrolary:rate} can be simplified as {\footnotesize $\mathcal{O}\left((1-p)\sigma^2L_p/(n\epsilon^2) \right)$} in this experimental setup. For $\hat{\zeta}_p^2$, the empirical result also matches the theory. These results show that in this experimental setup, adding shuffled synthetic data reduces both stochastic noise and function dissimilarity, and lead to a greater accuracy improvement and vast reduction on the number of rounds to reach the target accuracy. See Appendix~\ref{appendix_sec:convergence_param_in_dnn} for more information.

\begin{minipage}{0.55\textwidth}
\begin{table}[H]
\centering
\setlength{\tabcolsep}{2pt}
\begin{tabular}{llllc}
\toprule
 & Fedssyn & FedGEN & DENSE &$\frac{M_{\text{Fedssyn}}}{M_{\text{FedGEN}}}$\\  \midrule
CIFAR10 ($\alpha$=0.01) & \textbf{83.0 (4)} & 47.6 (70) & 15.2 (-) &6\% \\ 
CIFAR10 ($\alpha$=0.1) & \textbf{84.1 (8)} & 78.6 (17) & 40.1 (-) & 49\% \\ 
CIFAR100 ($\alpha$=0.01) & \textbf{52.0 (15)} & 35.1 (52) & 10.3 (-) & 29\%\\ 
CIFAR100 ($\alpha$=0.1) & \textbf{54.4 (19)} & 41.0 (84) & 13.5 (-) & 23\% \\\bottomrule 
\end{tabular}
\caption{Top-1 accuracy (number of rounds to reach target accuracy $m$ in parentheses), and the ratio between the memory cost to reach $m$. We obtain higher Top-1 accuracy and require lower memory cost compared to other synthetic data-based methods to reach accuracy $m$.}
\label{tab:comparison_with_other_methods}
\end{table}
\end{minipage}\hfill
\begin{minipage}{0.43\textwidth}
\begin{figure}[H]    
\vspace*{-1mm}
\includegraphics{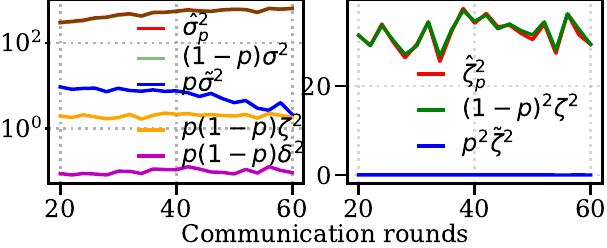}
\caption{The empirical observation of stochastic noise and gradient dissimilarity matches the theoretical statement (experiment using CIFAR10, 10 clients, $\alpha=0.1$, p=0.06)}
\label{fig:influence_of_sigma_zeta_main_text}
\end{figure}
\end{minipage}

\textbf{Practical implications} Our proposed framework Fedssyn is more suitable for applications such as disease diagnosis and explosives detection that usually require high precision and clients have enough computation resources seeing that Fedssyn can even reach the centralized accuracy in some of the high data heterogeneity scenarios and the saved communication cost is enormous (up to 95\%). Suppose the client is resource-constraint, a good strategy is to start with generating fewer synthetic images since shuffling even a small percentage (e.g. p=0.06) of the synthetic images into each client can already improve the Top-1 accuracy by 14\%-20\% (Fig.~\ref{fig:ablation_figure}) and a larger $p$ usually corresponds to a better accuracy until it saturates, e.g. reach the iid accuracy. Additionally, the client has the option of checking the synthetic images and only sharing the less sensitive synthetic images to the server to alleviate the information leakage. 

\textbf{Limitations} Training the generator locally may pose a computation requirement for each device. We can mitigate this issue with methods such as quantization~\citep{DBLP:journals/corr/Alistarh0TV16} and compression~\citep{https://doi.org/10.48550/arxiv.2002.11364, DBLP:journals/corr/abs-1901-09269}. Sending the synthetic images may also leak the training data distribution. One possible solution to improve the utility of Fedssyn is to make the generator differential private~\cite{yoon2021fedmix, yoon2018pategan, abadi2016deep}. We next provide a small-scale experiment that illustrates the performance of Fedssyn when the generator is differentially private given $\epsilon_{\text{DP}}=10$~\citep{abadi2016deep,dockhorn2022differentially}.

\section{Towards differentially private Fedssyn} 
We have shown the benefits of Fedssyn on top of multiple FL algorithms and across different datasets. However, sending synthetic data to the server may cause a certain level of privacy leakage. To address this issue, we train the data generator differentially private so that the shared synthetic images are privacy-friendly and are respecting the rigorous DP guarantee.

\begin{figure}[ht!]
    \centering
    \includegraphics{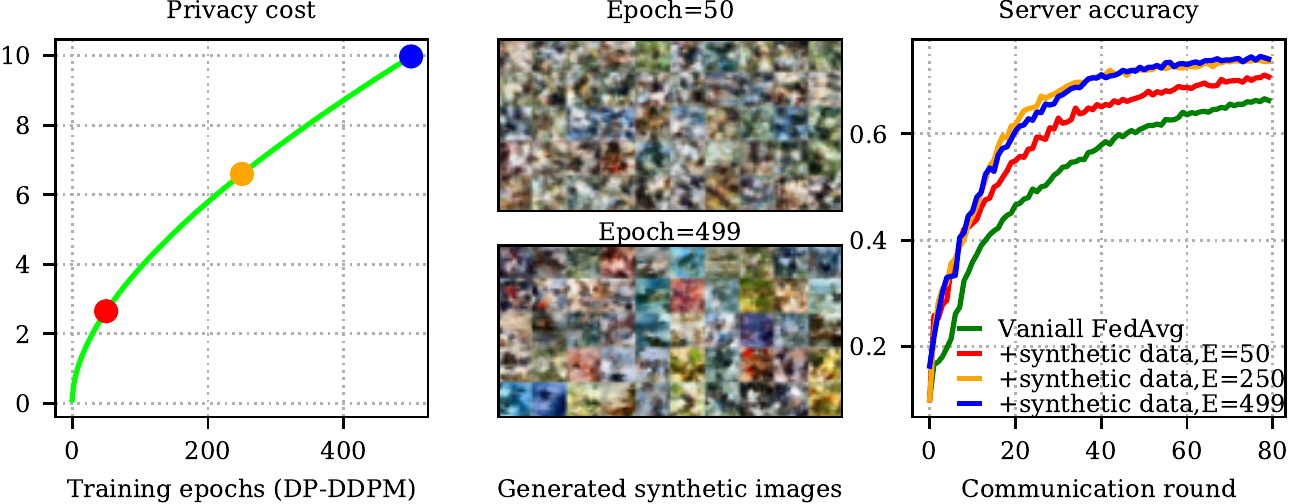}
    \caption{Performance of differentially private Fedssyn. From left to right, we show the privacy cost for training the differentially private DDPM, examples of synthetic images generated from two model checkpoints (annotated by dots in the left figure), and the corresponding federated learning performance. Shuffling differentially private synthetic images to each client can accelerate the convergence and improve the performance of vanilla FedAvg when the clients are heterogeneous in this experiment.}
    \label{fig:dp_aggregate_performance}
\end{figure}

Specifically, we train the diffusion model with rigorous DP guarantees based on DP-SGD~\citep{abadi2016deep} by clipping and adding noise to the gradients during training. Following~\cite{dockhorn2022differentially}, we train the DP-DDPM for 500 epochs under the private setting ($\epsilon_{\text{DP}}=10, \delta=1e-5$) with the noise multiplier as 1.906 and clipping threshold 1.0 based on Opacus~\citep{opacus} on each client. We use CIFAR10, ten clients with full client participation, and $\alpha=0.1$. We show the incurred privacy cost, examples of synthetic images, and the corresponding federated learning performance in Fig.~\ref{fig:dp_aggregate_performance}. See Fig.~\ref{fig:scaffold_dp_performance} for the performance of using SCAFFOLD as the baseline federated learning algorithm.

We observe that as the training of DP-DDPM progresses, the privacy cost increases. We then generate the synthetic images using three model checkpoints from different training stages (50 epochs, 250 epochs, and 499 epochs) on each client to illustrate the influence of the quality of the synthetic images on the performance of \texttt{Fedssyn}. Some example synthetic images are shown in the middle of Fig.~\ref{fig:dp_aggregate_performance}. We can see that the synthetic images are distorted and more privacy-friendly. We then aggregate, shuffle, and distribute these synthetic images back to the clients and train \texttt{Fedssyn} following the procedure described in section 4. We observe that shuffling privacy-friendly synthetic data, which are generated by the DDPM trained under the privacy budget $\epsilon_{\text{DP}}=10$, can accelerate convergence and improve accuracy compared to vanilla FedAvg. Different privacy budget and number of training images for training the DDPM can result in a different convergence behaviour. More extensive and fine grained experiments are needed to fully address this challenge. We leave this as our future work.


\section{Conclusion}
In this paper, we have rigorously analyzed the relation between data heterogeneity and the parameters in the convergence rate in FedAvg under standard stochastic noise and gradient dissimilarity assumptions. While previous works usually qualitatively represent data heterogeneity with gradient dissimilarity, we proposed a more quantifiable and precise characterization of how the changes in the data heterogeneity impact the parameters in the convergence rate using shuffling. 

Our work paves the way for a better understanding of the seemingly unreasonable effect of data heterogeneity in FL. The theoretical findings inspired us to present a more practical framework, which shuffles locally generated synthetic data, achieving nearly as good performance as centralized learning, even in some cases when the data heterogeneity is high. Our work reveals the importance of data distribution matching in FL, opening up future research directions. 

\section*{Reproducibility Statement}
We provide downloadable source code as part of the supplementary material. This code allows to reproduce our experimental evaluations show in the main part of the manuscript. The code for the additional verification on the dSprites dataset in Appendix~\ref{appendix_sec:dsprite} is not part of the submitted file. We will make this code and the genenerated datasets (such as in Figure~\ref{fig:dSprites_visualizations_from_checkpoints}) available on a public github repository. 

\section*{Acknowledgement}
Bo Li, Mikkel N. Schmidt, and Tommy S. Alstrøm thank for financial support from the European Union’s Horizon 2020 research and innovation programme under grant agreement no.\ 883390 (H2020-SU-SECU-2019 SERSing Project). 
Sebastian Stich thanks for partial financial support from a Meta Privacy Enhancing Technologies Research Award 2022.
Bo Li thanks for the financial support from the Otto Mønsted Foundation and appreciates the discussion with Xiaowen Jiang.


\bibliography{iclr2024_conference}
\bibliographystyle{tmlr}

\newpage 
\appendix
\numberwithin{equation}{section}
\numberwithin{figure}{section}
\numberwithin{table}{section}

\section{Appendix}

We first provide the proof for the lemma. We then show the comparison with other synthetic-data based works. Following that, we show the extra experimental details and experimental results on two other dataset MNIST and dSprites. 

\subsection{Proof}
\label{appendix_sec:lemma}

We study the distributed stochastic optimization problem: 
\[f(\xx,\mathcal{D}) := \mathbb{E}_{\xi\sim\mathcal{D}}[F(\xx,\xi)], \quad f(\xx,\mathcal{D}_i) := \mathbb{E}_{\xi\sim\mathcal{D}_i}[ F(\xx,\xi)]\] 
where $\xi$ can be a random data point or a random function, $\mathcal{D}_i$ denotes the distribution of $\xi$ on client $i$ and $\mathcal{D}$ denotes the distribution of $\xi$ over all the clients. 

We argue that it is possibly not be feasible to assume access to the distribution $\mathcal{D}$ as collecting data from clients goes against the goal of protecting privacy in federated learning. Instead, we use an approximation (e.g. synthetic data) $\tilde{\mathcal{D}}$ for carrying out the proof. We will demonstrate the impact of the difference between $\mathcal{D}$ and $\tilde{\mathcal{D}}$ in the following sections. 

Given the effective data $\mathcal{D}_i\cup\tilde{\mathcal{D}}$ on each client we can formulate the optimization problem as:
\begin{equation}
    F(\xx, \xi\sim\mathcal{D}_i\cup\tilde{\mathcal{D}};b) := \left.\begin{cases}
        F(\xx,\xi\sim\mathcal{D}_i\mid b=0) & \quad \text{with probability } 1-p \\ 
        F(\xx,\xi\sim\tilde{\mathcal{D}}\mid b=1) &\quad \text{with probability } p \\ 
    \end{cases} \right. 
    \label{eq:def_F_i}
\end{equation}
where $b$ is a random variable that indicates where a data point is drawn from. When $b=0$ (with probability $1-p$), we assume that we draw from a worker's own data $\mathcal{D}_i$. When $b=1$ (with probability $p$), we assume that we draw from the uniform distribution $\tilde{\mathcal{D}}$. Take the expectation with respect to $b$, we have:
\begin{subequations}
    \begin{equation}
        F(\xx, \xi\sim\mathcal{D}_i\cup\tilde{\mathcal{D}}) = \mathbb{E}_b[F(\xx, \xi;b)] = (1-p)F(\xx, \xi\sim\mathcal{D}_i) + pF(\xx, \xi\sim\tilde{\mathcal{D}})
    \end{equation}
    \begin{equation}
        \begin{split}
            f(\xx, \mathcal{D}_i\cup\tilde{\mathcal{D}}) &= \mathbb{E}_{\xi\sim\mathcal{D}_i\cup\tilde{\mathcal{D}}}[F(\xx, \xi)] = (1-p)\mathbb{E}_{\xi\sim\mathcal{D}_i}[F(\xx, \xi)] + p\mathbb{E}_{\xi\sim\tilde{\mathcal{D}}}[F(\xx, \xi)]\\
            &= (1-p)f(\xx,\mathcal{D}_i) + pf(\xx,\tilde{\mathcal{D}})            
        \end{split}
    \end{equation}
    \begin{equation}
        \begin{split}
            f(\xx,\mathcal{D}\cup\tilde{\mathcal{D}}) &= \mathbb{E}_{\xi\sim\mathcal{D}\cup\tilde{\mathcal{D}}}[F(\xx, \xi)] = (1-p)\mathbb{E}_{\xi\sim\mathcal{D}}[F(\xx, \xi)] + p\mathbb{E}_{\xi\sim\tilde{\mathcal{D}}}[F(\xx, \xi)] \\
            &= (1-p)f(\xx,\mathcal{D}) + pf(\xx,\tilde{\mathcal{D}})
        \end{split}
    \end{equation}
    \label{eq:full_definitions}
\end{subequations}

\begin{lemmap}{A-1}[variance with probability]
If the random variable $X$ is discrete with each element $x_i$ appearing with a probability $p_i$ such that $p(x_i) = p_i$, then:
\begin{equation*}
    \mathbb{E}||X - \mu||^2 = \sum_{i=1}^n p_i||x_i - \mu||^2, \quad \mu := \frac{1}{n}\sum_{i=1}^np_ix_i \,.
\end{equation*}
\label{lemma:variance_with_probability}
\end{lemmap}

\begin{lemmap}{A-2}[shuffled gradient dissimilarity]
If Assumption 1 holds, then $\forall \xx\in\mathbb{R}^d$:
\begin{equation*}
    \hat{\zeta}_p^2 := \mathbb{E}||\nabla f(\xx, \mathcal{D}_i\cup\tilde{\mathcal{D}}) - \nabla f(\xx,\mathcal{D}\cup\tilde{\mathcal{D}})||^2 \leq (1-p)^2\zeta^2 \,,
\end{equation*}
where here the expectation is taken over a random index $i \in [N]$.
\begin{proof}

Given the definition from Eq.~\ref{eq:full_definitions}, we have:
\begin{equation}
    \begin{split}
        \hat{\zeta}^2_p :&= \mathbb{E}||\nabla f(\xx, \mathcal{D}_i\cup\tilde{\mathcal{D}}) - \nabla f(\xx, \mathcal{D}\cup\tilde{\mathcal{D}})||^2 \\
        &= \mathbb{E}||(1-p)\nabla f(\xx, \mathcal{D}_i) + p\nabla f(\xx,\tilde{\mathcal{D}}) - (1-p)\nabla f(\xx,\mathcal{D}) - p\nabla f(\xx,\tilde{\mathcal{D}})||^2 \\
        &= (1-p)^2\mathbb{E}||\nabla f(\xx,\mathcal{D}_i) - \nabla f(\xx,\mathcal{D})||^2 \\
        &\leq (1-p)^2\zeta^2\,.
    \end{split}
\end{equation}

\end{proof}
\end{lemmap}

\begin{lemmap}{A-3}[shuffled stochastic noise]
If Assumption~\ref{assum:original_gradient_dissimilarity},~\ref{assump:stochastic_noise}, and~\ref{assum:delta} hold and let $\tilde{\sigma}^2_{\text{avg}}$ being the stochastic noise from using $\tilde{\mathcal{D}}$ such that $\mathbb{E}_{\xi\sim\tilde{\mathcal{D}}}||\nabla F(\xx, \xi) - f(\xx,\tilde{\mathcal{D}})||^2 \leq \tilde{\sigma}_{\text{avg}}^2$, then $\forall \xx\in\mathbb{R}^d$, we have:
\[\hat{\sigma}_p^2 := \mathbb{E}_{\xi\sim\mathcal{D}_i\cup\tilde{\mathcal{D}}}||\nabla F(\xx, \xi) - \nabla f(\xx,\mathcal{D}_i\cup\tilde{\mathcal{D}})||^2 \leq (1-p)\sigma^2+p\tilde{\sigma}^2_{\text{avg}} + p(1-p)||\nabla f(\xx,\mathcal{D}_i) - \nabla f(\xx,\tilde{\mathcal{D}})||^2\]%
\[\mathbb{E}[\hat{\sigma}^2_p] \leq (1-p)\sigma^2+p\tilde{\sigma}^2_{\text{avg}} + p(1-p)\zeta^2+p(1-p)\delta^2\]
\end{lemmap}
\begin{proof}
Based on the definition in Eq.~\ref{eq:full_definitions}, we have: 
\begin{equation*}
    \begin{split}
        \hat{\sigma}_p^2 :&= \mathbb{E}_{\xi\sim\mathcal{D}_i\cup\tilde{\mathcal{D}}}||\nabla F(\xx, \xi) - \nabla f(\xx, \mathcal{D}_i\cup\tilde{\mathcal{D}})||^2 \\
        &= (1-p)\underbrace{\mathbb{E}_{\xi\sim\mathcal{D}_i}||\nabla F(\xx, \xi) - \nabla f(\xx,\mathcal{D}_i\cup\tilde{\mathcal{D}})||^2}_{\mathcal{A}_1} + p\underbrace{\mathbb{E}_{\xi\sim\tilde{\mathcal{D}}}||\nabla F(\xx, \xi) - \nabla f(\xx,\mathcal{D}_i\cup\tilde{\mathcal{D}})||^2}_{\mathcal{A}_2}
    \end{split}
\end{equation*}
We use Lemma~\ref{lemma:variance_with_probability} in the last equality. We next give the bound for $\mathcal{A}_1$ and $\mathcal{A}_2$. 
\begin{equation*}
    \begin{split}
        \mathcal{A}_1 :&= \mathbb{E}_{\xi\sim\mathcal{D}_i}||\nabla F(\xx, \xi) - \nabla f(\xx, \mathcal{D}_i) + \nabla f(\xx, \mathcal{D}_i) - \nabla f(\xx,\mathcal{D}_i\cup\tilde{\mathcal{D}})||^2 \\
        &= \mathbb{E}_{\xi\sim\mathcal{D}_i}||\nabla F(\xx,\xi) - \nabla f(\xx,\mathcal{D}_i)||^2 + ||\nabla f(\xx,\mathcal{D}_i) - \nabla f(\xx,\mathcal{D}_i\cup\tilde{\mathcal{D}})||^2 \\
        &\leq \sigma^2 + ||\nabla f(\xx,\mathcal{D}_i) - (1-p)f(\xx,\mathcal{D}_i) - pf(\xx,\tilde{\mathcal{D}})||^2 \\
        &\leq \sigma^2 + p^2||\nabla f(\xx,\mathcal{D}_i) - \nabla f(\xx,\tilde{\mathcal{D}})||^2 \\
    \end{split}
\end{equation*}
Similarly, we can bound $\mathcal{A}_2$ as:
\begin{equation*}
    \begin{split}
        \mathcal{A}_2 :&= \mathbb{E}_{\xi\sim\tilde{\mathcal{D}}}||\nabla F(\xx, \xi) - \nabla f(\xx,\mathcal{D}_i\cup\tilde{\mathcal{D}})||^2 \\
        &= \mathbb{E}_{\xi\sim\tilde{\mathcal{D}}}||\nabla F(\xx, \xi) - \nabla f(\xx,\tilde{\mathcal{D}})||^2 + ||\nabla f(\xx,\tilde{\mathcal{D}}) - \nabla f(\xx,\mathcal{D}_i\cup\tilde{\mathcal{D}})||^2 \\ 
        &\leq \tilde{\sigma}^2_{\text{avg}} + ||\nabla f(\xx,\tilde{\mathcal{D}}) - (1-p)f(\xx,\mathcal{D}_i) - pf(\xx,\tilde{\mathcal{D}})||^2 \\
        &= \tilde{\sigma}^2_{\text{avg}} + (1-p)^2||\nabla f(\xx,\mathcal{D}_i) - \nabla f(\xx,\tilde{\mathcal{D}})||^2 
    \end{split}
\end{equation*}
Taking the bound for $\mathcal{A}_1$ and $\mathcal{A}_2$ back to the expression of $\hat{\sigma}_p^2$, we have:
\begin{equation*}
    \hat{\sigma}_p^2 \leq (1-p)\sigma^2+p\tilde{\sigma}^2_{\text{avg}} + p(1-p)||\nabla f(\xx,\mathcal{D}_i) - \nabla f(\xx,\tilde{\mathcal{D}})||^2 
\end{equation*}

If we take the expectation w.r.t the clients for both sides, we then have:
\begin{equation}
    \begin{split}
        \mathbb{E}[\hat{\sigma}_p^2] &\leq (1-p)\sigma^2+p\tilde{\sigma}^2_{\text{avg}} + p(1-p)\mathbb{E}||\nabla f(\xx,\mathcal{D}_i) - \nabla f(\xx, \tilde{\mathcal{D}})||^2 \\    
        &= (1-p)\sigma^2+p\tilde{\sigma}^2_{\text{avg}} + p(1-p)\mathbb{E}||\nabla f(\xx,\mathcal{D}_i) - \nabla f(\xx,\mathcal{D}) + \nabla f(\xx,\mathcal{D}) - \nabla f(\xx,\tilde{\mathcal{D}})||^2 \\
        &= (1-p)\sigma^2+p\tilde{\sigma}^2_{\text{avg}} + p(1-p)\mathbb{E}||\nabla f(\xx,\mathcal{D}_i) - \nabla f(\xx,\mathcal{D})||^2 + p(1-p)||\nabla f(\xx,\mathcal{D}) - \nabla f(\xx,\tilde{\mathcal{D}})||^2 \\
        &\leq (1-p)\sigma^2+p\tilde{\sigma}^2_{\text{avg}} + p(1-p)\zeta^2 + p(1-p)||\nabla f(\xx,\mathcal{D}) - \nabla f(\xx,\tilde{\mathcal{D}})||^2 \\
        &\leq (1-p)\sigma^2+p\tilde{\sigma}^2_{\text{avg}} + p(1-p)\zeta^2 + p(1-p)\delta^2 
    \end{split}
\end{equation}

The inequalities use the Assumption~\ref{assum:original_gradient_dissimilarity} and~\ref{assum:delta}. If $\tilde{\mathcal{D}} = \mathcal{D}$, then $\delta^2 = 0$ by definition. We can then rewrite the bound for the stochastic noise as $\mathbb{E}[\hat{\sigma}_p^2] \leq (1-p)\sigma^2+p\sigma^2_{\text{avg}} + p(1-p)\zeta^2$
\end{proof}

\begin{lemmap}{A-4}[shuffled smoothness]
Let $L_{\max}$ and $L_{\text{avg}}$ denote the maximum and average smoothness, i.e.\ it holds
\[
 \left\| \nabla f(\xx,\mathcal{D}_i)- \nabla f (\yy,\mathcal{D}_i) \right\| \leq L_{\max} \left\| \xx - \yy \right\| \,, \qquad \forall \xx ,\yy \in \mathbb{R}^d\,, i \in [n]\,,
\]
and
\[
\left\| \nabla f(\xx,\mathcal{D})- \nabla f(\yy,\mathcal{D}) \right\| \leq L_{\text{avg}} \left\| \xx - \yy \right\| \,, \qquad \forall \xx ,\yy \in \mathbb{R}^d\,.
\]
and
\[\left\|\nabla f(\xx, \tilde{\mathcal{D}}) - \nabla f(\yy, \tilde{\mathcal{D}})\right\| \leq \tilde{L}_{\text{avg}}\left\|\xx - \yy\right\| \,, \qquad  \forall \xx, \yy\in\mathbb{R}^d\]
Then we have:
\begin{equation*}
    \mathbb{E}L_p \leq (1-p)L_{\max}+p\tilde{L}_{\text{avg}} \,,
\end{equation*}    
where $\mathbb{E}L_p$ denotes an upper bound on the smoothness constant of $f(\xx,\mathcal{D}_i\cup\tilde{\mathcal{D}})$.
\end{lemmap}
\begin{proof}
Given the client index $i$, by definition, we have $f(\xx,\mathcal{D}_i\cup\tilde{\mathcal{D}}) = (1-p)f(\xx,\mathcal{D}_i) + pf(\xx;\tilde{\mathcal{D}})$,
\begin{align*}
    || \nabla f(\xx,\mathcal{D}_i\cup\tilde{\mathcal{D}}) - \nabla f(\yy,\mathcal{D}_i\cup\tilde{\mathcal{D}}) || &= || (1-p)f(\xx,\mathcal{D}_i) + pf(\xx;\tilde{\mathcal{D}}) - \nabla f(\yy,\mathcal{D}_i\cup\tilde{\mathcal{D}})|| \\
    &= ||(1-p)\left(\nabla f(\xx,\mathcal{D}_i) - \nabla f(\yy,\mathcal{D}_i) \right) + p\left(\nabla f(\xx;\tilde{\mathcal{D}}) - \nabla f(\yy;\tilde{\mathcal{D}})\right)|| \\
    &\leq (1-p)||\nabla f(\xx,\mathcal{D}_i) - \nabla f(\yy,\mathcal{D}_i)|| + p||\nabla f(\xx;\tilde{\mathcal{D}}) - \nabla f(\yy;\tilde{\mathcal{D}})|| \\
    &\leq \left( (1-p)L_{\max} + pL_{\text{avg}}\right)||\xx-\yy||\,.
\end{align*}
\end{proof}

\subsection{Extra related work comparison}\label{appendix_sec:related_work_comparison}

\textbf{Experimental details for replicating the FedGEN and DENSE results} For FedGEN, we follow the default hyperparameter setup as described in the paper~\cite{DBLP:journals/corr/abs-2105-10056} and the official repository. We tune the optimal stepsize from \{0.05, 0.1, 0.2\}. We run FedGEN for 100 communication rounds and report the final Top-1 accuracy using the server model on the test dataset. For DENSE, we follow the hyperparameter setup as described in the paper~\cite{zhang2022dense} and the official GitHub repository. We have tried our best to improve the quality of the server model obtained after one round of communication as have done in~\cite{zhang2022dense}. However, we still observe that such server model is not sufficiently good for the global distillation learning, especially when the data heterogeneity is high ($\alpha=0.01$). Therefore, we performed ten rounds of communication and then used the server model for global distillation learning. We can possibly improve the performance of FedGEN and DENSE with more careful tuning of the random seeds and other hyperparameters. 

\begin{figure}[ht!]
    \centering
    \includegraphics{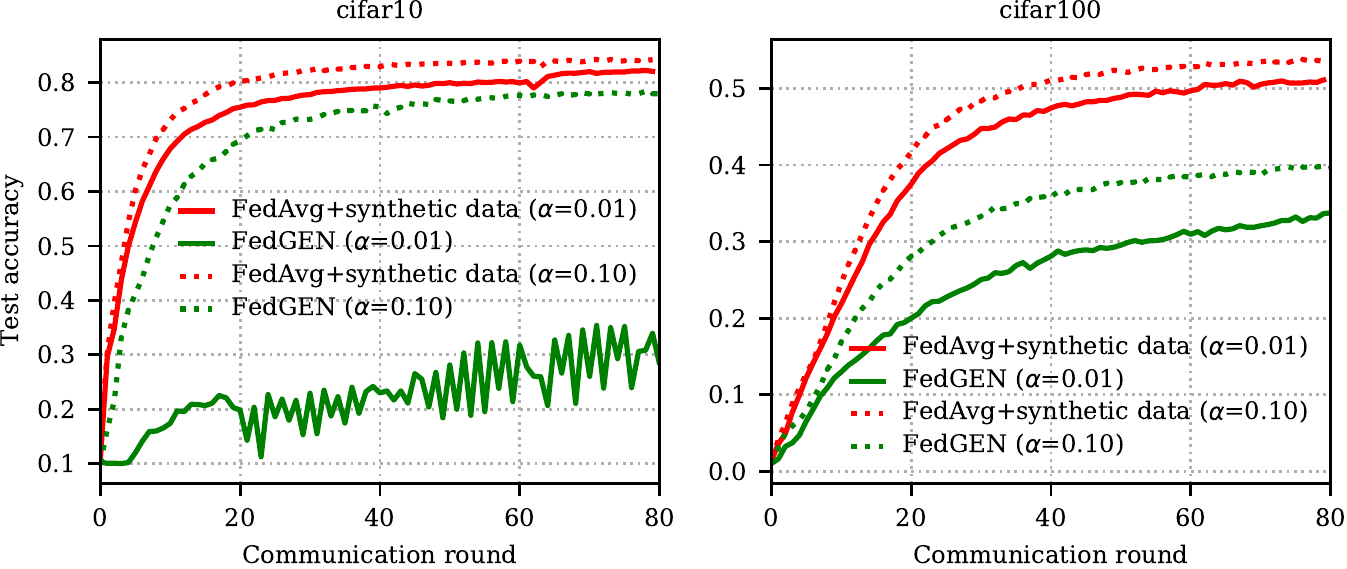}
    \caption{Test accuracy for two methods FedAvg + synthetic data and FedGEN.Comparably, our proposed method reaches to a better test accuracy faster than FedGEN. The unstable performance of when $\alpha=0.01$ on CIFAR10 dataset using FedGEN can be potentially improved with more careful hyperparameter tuning.}
    \label{fig:test_accuracy_for_fed_gen}
\end{figure}

\textbf{Experimental details for replicating the FedDM results} For FedDM~\cite{xiong2023feddm}, we follow the hyperparameter setup as described in the paper~\cite{xiong2023feddm} and also the downloadable supplementary file from\footnote{https://openreview.net/forum?id=40RNVzSoCqD}. We choose the non-differentially private FedDM with ten clients. We run the experiment on the CIFAR10 dataset with $\alpha$ being 0.01 and 0.1. We use the same classification model VGG-11 as in our experiment. We run each experiment two times with different initializations and report the averaged test accuracy. Note that we can possibly improve the performance of FedDM using a different classification model. 

\begin{figure}[ht!]
    \centering
    \includegraphics{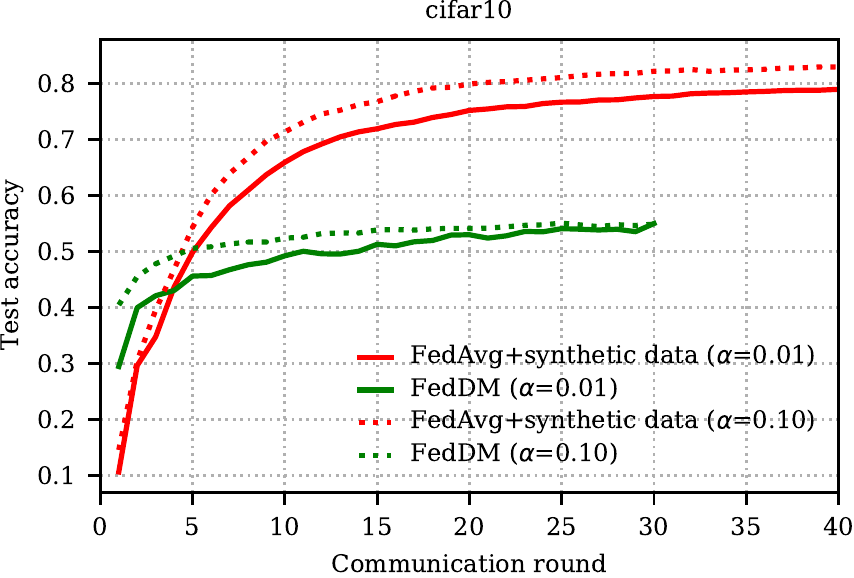}
    \caption{Test accuracy for two methods FedAvg + synthetic data and FedDM. Comparably, FedDM performs better than our approach at the beginning. However, as the communication round increases, our method performs significantly better than FedDM. FedDM is more robust against different levels of data heterogeneity, seeing that the test accuracy gap between different levels of data heterogeneity is small.}
    \label{fig:test_accuracy_for_fed_dm}
\end{figure}

In Fig.~\ref{fig:test_accuracy_for_fed_dm}, we observe that FedDM initially outperforms our approach. However, as training progresses, our method demonstrates a significant improvement over FedDM. Additionally, FedDM seems more robust against different levels of data heterogeneity as the test accuracy gap between when $\alpha=0.01$ and $\alpha=0.1$ is small. Comparably, our method needs to communicate 30,070 MB of parameters (over 40 rounds), whereas FedDM only needs to communicate 22,508 MB of parameters (over 30 rounds, with each round transmitting around 750 MB of parameters) between the client and the server to achieve a stabilized test accuracy.

\subsection{Extra experimental details}

\begin{table}[ht!]
\caption{Extension of Table~\ref{tab:summary_result}. The required number of communication rounds for Fedssyn (for vanilla FL algorithms) to reach the target accuracy $m$. $m$ is chosen so that most FL algorithms can get such accuracy within 100 communication rounds. The numbers in the parentheses represent the number of rounds required for the vanilla FL algorithms to reach $m$-accuracy. }
\label{appendix_table:extension_of_the_main_table}
\resizebox{0.95\textwidth}{!}
{\begin{minipage}{\textwidth}
\begin{tabular}{lllllllll}
\toprule
              & \multicolumn{2}{c}{Full participation}              & \multicolumn{6}{c}{Partial participation} \\ 
              \cmidrule(lr){2-3}\cmidrule(l){4-9} & \multicolumn{2}{c}{N=10} &  \multicolumn{2}{c}{N=10 (C=40\%)} & \multicolumn{2}{c}{N=40 (C=20\%)} & \multicolumn{2}{c}{N=100 (C=10\%)} \\\cmidrule(lr){2-3}\cmidrule(l){4-5}\cmidrule(lr){6-7}\cmidrule(l){8-9}
 &$\alpha=0.01$ & $\alpha=0.1$ & $\alpha=0.01$  & $\alpha=0.1$   & $\alpha=0.01$           & $\alpha=0.1$ & $\alpha=0.01$           & $\alpha=0.1$           \\ \cmidrule(lr){2-9} 
 & \multicolumn{8}{c}{CIFAR10} \\ \midrule 
 & m=44\% & m=66\% & m=44\% & m=66\% & m=44\% & m=66\% & m=44\% & m=66\% \\ \midrule
FedAvg & 4 (88) & 8 (68) & 4 ($>$100) & 8 (45) & 12 (100) & 21 (59) & 19 (93) & 38 (>100) \\
Scaffold & 4 (37) & 7 (43) & 5 (>100) & 10 (26) & 12 (82) & 20 (54) & 19 (>100)& 31 (84) \\
FedProx & 4 (94) & 8 (43) & 4 (>100) & 9 (74) & 9 (>100) & 15 (71) & 31 (>100) & 37 (>100) \\
FedDyn & 3 (49) & 5 (74) & 4 (67) & 7 (75) & 7 (>100) & 12 (>100) & 17 (>100) & 31 (>100) \\
FedPVR & 3 (58) & 7 (30) & 4 (>100) & 8 (76) & 10 (85) & 21 (59) & 21 (>100) & 35 (>100)\\\midrule
 & \multicolumn{8}{c}{CIFAR100} \\ \midrule 
 & m=30\% & m=40\% & m=20\% & m=20\% & m=30\% & m=40\% & m=30\% & m=30\% \\ \midrule
FedAvg & 15 (89) & 19 (>100) & 17 (>100) & 21 (>100) & 37 (>100) & 89 (>100) & 70 (>100) & 59 (>100) \\
Scaffold & 14 (41) & 19 (55) & 15 (54) & 21 (100) & 24 (>100) & 71 (>100) & 61 (100) & 52 (>100) \\
FedProx & 17 (100) & 23 (99) & 17 (>100)& 23 (>100)& 33 (>100)& >100 (-) & 75 (>100) & 60 (>100) \\
FedDyn & 10 (36) & 12 (66) & 13 (>100) & 15 (100)& 25 (>100)& 78 (>100) & 61 (>100) & 60 (>100) \\
FedPVR & 14 (85) & 19 (38) & 15 (>100)& 20 (100)& 23 (>100)& 59 (>100) & 71 (>100) & 58 (>100) \\\bottomrule
\end{tabular}
\end{minipage}}
\end{table}

\subsubsection{Experimental results on MNIST dataset}\label{appendix_sec:mnist}
We consider a distributed multi-class classification problem on MNIST dataset given $(A_i, \boldsymbol{b}_i)$ as the dataset on worker $i$ with:
\begin{subequations}
\begin{equation}
    \yy_i = \text{softmax}(A_i\xx), \quad A_i\sim \mathbb{R}^{n_i\times 784}, \xx\in \mathbb{R}^{784\times10}, \yy_i\in \mathbb{R}^{n_i\times 10}
\end{equation}
\begin{equation}
    f_i(\xx) := \frac{1}{n_i}\sum_{j=1}^{n_i}\frac{1}{10}\sum_{k=1}^{10} (\boldsymbol{b}_{ijk}\log(\yy_{ijk}))
\end{equation}
\end{subequations}

We construct the training dataset by subsampling 1024 images from each class from the MNIST training dataset. We use the test dataset to evaluate the performance of the server model (10000 images). We consider two scenarios for splitting the dataset across workers: 1) \texttt{iid}, where each worker has a similar number of images across classes 2) \texttt{split by class}, where each worker can only see images from a single class. We set 10 workers, and each worker has 1024 images $n_i=1024$. We experiment with a different number of local epochs $E$ on each worker and shuffling percentage $p$. For a given $p$, we subsample $n_ip$ data points randomly from each worker and allocate them in $\tilde{\mathcal{D}}$. We then shuffle $\tilde{\mathcal{D}}$ and distribute it equally to all workers such that each worker has $n_ip$ shuffled data points and $n_i(1-p)$ local data points. We choose to control the stochastic noise $\bar{\sigma}^2$ by adding Gaussian noise to every gradient:
\begin{equation}
    \yy_i \leftarrow \yy_i - \eta (\nabla f_i(\yy_i) + s), \quad s\sim \mathcal{N}(0, \frac{\bar{\sigma}^2}{784})
\end{equation}

\begin{figure}[ht!]
    \centering
    \includegraphics[width=\textwidth]{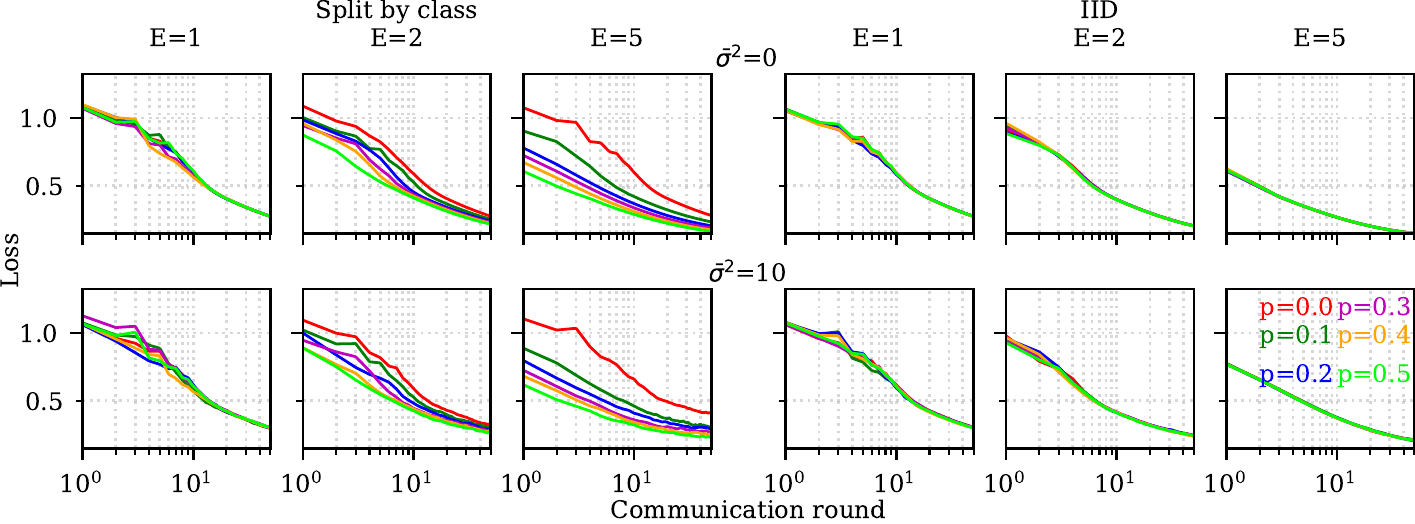}
    \caption{Test loss using the server model over communication rounds. When the dataset is IID across workers, adding shuffled dataset does not influence the learning speed. However, when workers have heterogeneous datasets (\texttt{split by class}), shuffling a small percentage of the dataset can improve the convergence speed, especially when the number of local epochs is high (e.g., $E=5$)}
    \label{fig:synthetic_mnist_example}
\end{figure}

Fig.~\ref{fig:synthetic_mnist_example} shows the results. When the dataset is IID across workers, adding the shuffled dataset to each worker does not influence the learning speed. This agrees with the lemmas that we have shown in the main text. When the dataset is non-iid across workers, shuffling a small percentage of the local dataset can highly improve the convergence, especially when the number of local epochs is high, e.g., $E=5$. We can achieve a slightly better speedup when the stochastic noise is non-zero, which is most of the cases in the neural network training where we use mini-batch SGD.

\begin{figure}[ht!]
    \centering
    \includegraphics{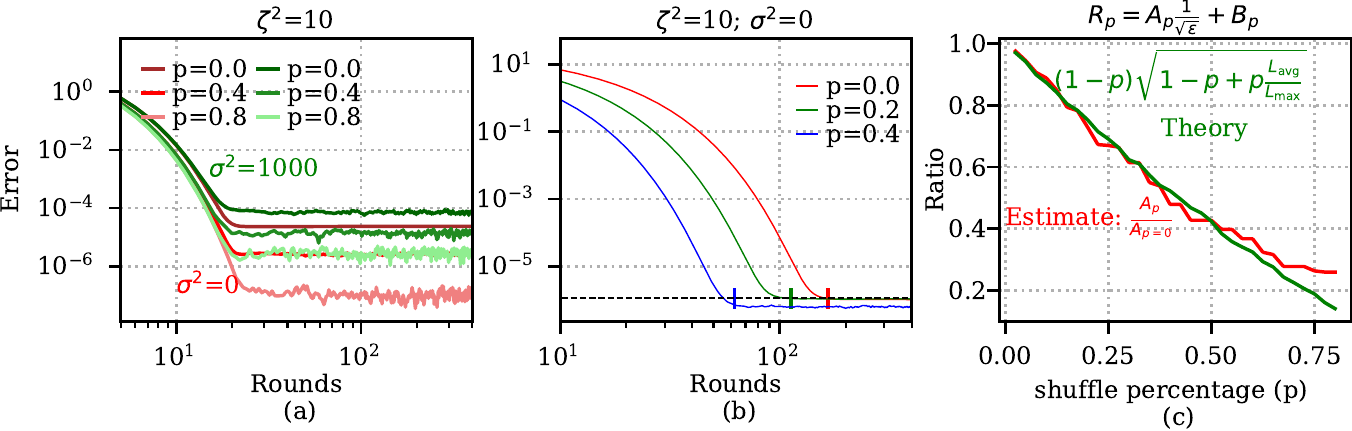}%
    \caption{Convergence of $\frac{1}{n}\sum_{i=1}^n||\xx_i^t-\xx^\star||^2$. (a) With a fixed $\zeta^2$ and step size, shuffling reduces the optimal error more when the stochastic noise is low (b) When gradient dissimilarity $\zeta^2$ dominates the convergence, we obtain a super-linear speedup in the number of rounds to reach $\epsilon$ by shuffling more data. The vertical bar shows the theoretical number of rounds to reach $\epsilon$ (c) The estimated ratio matches the theoretical ratio when the shuffle percentage is small (e.g., $p\leq 0.6$). The term {\scriptsize $\sqrt{1-p+p\frac{L_{\text{avg}}}{L_\text{max}}}$} comes from the Lipschitz constant ratio {\scriptsize $\sqrt{\frac{L_{p}}{L_{p=0}}}$}.}%
    \label{fig:synthetic_exp}
\end{figure}

\subsubsection{Extra explanation for Fig.2}\label{appendix_sec:extra_explanation_for_fig2}
We consider a distributed least squares objective $f(\xx) := \frac{1}{n}\sum_{i=1}^n\bigl[f_i(\xx):=\frac{1}{2n_i}\sum_{j=1}^{n_i}||\boldsymbol{A}_{ij}\xx - \boldsymbol{b}_{ij}||^2\bigr]$ with $\boldsymbol A_{ij}=i\boldsymbol I_d$, 
$\boldsymbol{\mu}_{i}\sim \mathcal{N}(0, \zeta^2 (id)^{-2}\boldsymbol{I}_d)$, and $\boldsymbol{b}_{ij}\sim \mathcal{N}(\boldsymbol{\mu}_i, \sigma^2 (id)^{-2}\boldsymbol{I}_d)$, where $\zeta^2$ controls the function similarity and $\sigma^2$ controls the stochastic noise (matching parameters in Corollary I). On each worker, we generate $n_i=100$ pairs of $\{\boldsymbol{A}_{ij}, \boldsymbol{b}_{ij}\}$ with $d=25$. We randomly sample $p\cdot n_i$ pairs of $\{\boldsymbol{A}_{ij}, \boldsymbol{b}_{ij}\}$ from all the clients. We then aggregate, shuffle, and redistribute them equally to each client such that the number of functions per client is still $n_i$. We depict the influence of shuffling on different parameters in Fig.~\ref{fig:synthetic_exp}.
To concisely match out theory, we need to consider that also the curvature of the function changes and measure  $L_p$,
$L_{\text{max}}:=\max_i \big\|\frac{1}{n_i}\sum_{j=1}^{n_i}\boldsymbol{A}_{ij}^T\boldsymbol{A}_{ij}\big\|$ and $L_{\text{avg}}:=\frac{1}{n}\sum_{i=1}^n\big\|\frac{1}{n_i}\sum_{j=1}^{n_i}\boldsymbol{A}_{ij}^T\boldsymbol{A}_{ij}\|$.

Fig.~\ref{fig:synthetic_exp} (a) shows the influence of shuffling when we vary the stochastic noise given a fixed $\zeta^2$ and stepsize. We observe that in the high-noise regime, shuffling gives a smaller reduction on the optimal error than when $\sigma^2=0$ as {\footnotesize $\mathcal{O}\left(\frac{(1-p)\sigma^2+p\sigma^2_{\text{avg}}}{\epsilon}\right)$} tends to dominate no matter how much data we shuffle. Fig.~\ref{fig:synthetic_exp} (b) shows the influence of shuffling when $\sigma^2=0$ and {\footnotesize $\mathcal{O}\left(\frac{\sqrt{L_p}\tau(1-p)\zeta}{\sqrt{\epsilon}}\right)$} dominates the convergence. We tune the step size for each experiment to reach the target accuracy ({\footnotesize $\epsilon=1.1\cdot10^{-6}$}) with the fewest rounds. The theoretical number of rounds required to reach $\epsilon$, indicated by the vertical bars, were determined by calculating {\footnotesize $R_p=\frac{A_p}{\sqrt{\epsilon}}+B_p$}, where $A_p := \frac{\sqrt{L_p}\tau\hat{\zeta}_p}{\mu}$ is the coefficient that depends on $\zeta^2$ and $L_p$. To estimate $A_p$, we measure the number of rounds to reach different accuracies and fit a linear line between $R_p$ and $\frac{1}{\sqrt{\epsilon}}$. Fig.~\ref{fig:synthetic_exp} (b) shows that the empirical speedup matches the theoretical speedup as the observed and theoretical number of rounds to reach the target accuracy are very close. For better visualization, we compare the theoretical and estimated ratio in Fig.~\ref{fig:synthetic_exp} (c) given $\sigma^2=0$. Fig.~\ref{fig:synthetic_exp}(c) shows the ratio between the coefficient $\frac{A_p}{A_{p=0}}$ which boils down to {\footnotesize $\mathcal{O}\left(\frac{\sqrt{L_{p}}\zeta_p}{\sqrt{L_{p=0}}\zeta_{p=0}}\right)$}. The nearly matching lines in Fig.~\ref{fig:synthetic_exp} (c), especially when $p$ is small, verify our theoretical statement about the impact of data heterogeneity on the convergence parameters. When $p$ is large, $\sigma^2$ and the constant term tend to dominate the convergence. A linear line between $R_p$ and $\frac{1}{\sqrt{\epsilon}}$ is incomplete to explain the convergence rate.

\subsubsection{Communication cost}
We calculate the number of transmitted parameters when FedAvg is used in Table~\ref{appendix_table:communication_cost}, which shows that we can reduce the number of transmitted parameters up to 95\% to achieve a certain level of accuracy. For methods such as SCAFFOLD~\cite{DBLP:journals/corr/abs-1910-06378} and FedPVR~\cite{DBLP:journals/corr/abs-2212-02191}, we also need to take into account the transmitted control variates when we calculate the number of transmitted parameters per round. 

\begin{table}[ht!]
\setlength{\tabcolsep}{2pt}
\caption{The saved communication cost of using Fedssyn compared to the vanilla FedAvg (percentage between the number of transmitted parameters). $M_s$ is the size of the transmitted synthetic data and $M_c$ is the size of the FL model. $M_c$ in our experiment is around 37.2 MB. Fedssyn transmits significantly less number of parameters compared to the vanilla FedAvg to reach the same level of accuracy.}\label{appendix_table:communication_cost}
\begin{center}
\begin{tabular}{lllll}
\toprule
  Number of client (participation rate)   & \multicolumn{2}{c}{CIFAR10} & \multicolumn{2}{c}{CIFAR100} \\ \toprule
     &  $\alpha=0.01$  &  $\alpha=0.1$    & $\alpha=0.01$  &  $\alpha=0.1$    \\ \midrule
N=10 (C=1.0) ($M_s=15.5$MB) &  95.0\% & 87.6\%  & 82.5\%   & $>$80.2\%  \\
N=10 (C=0.4) ($M_s=15.5$MB) & $>$95.6\% & 80.9\% & $>$82.6\% & $>$78.6\% \\
N=40 (C=0.2) ($M_s=0.39$MB)&  87.9\% & 63.8\%  & $>$62.6\% & $>$8.2\% \\ 
N=100 (C=0.1) ($M_s=0.16$MB) & 79.6\% & $>$61.2\% & $>$28.6\% & $>$41.0\% \\ \bottomrule
\end{tabular}    
\end{center}
\end{table}

\begin{figure}[ht!]
    \centering
    \includegraphics{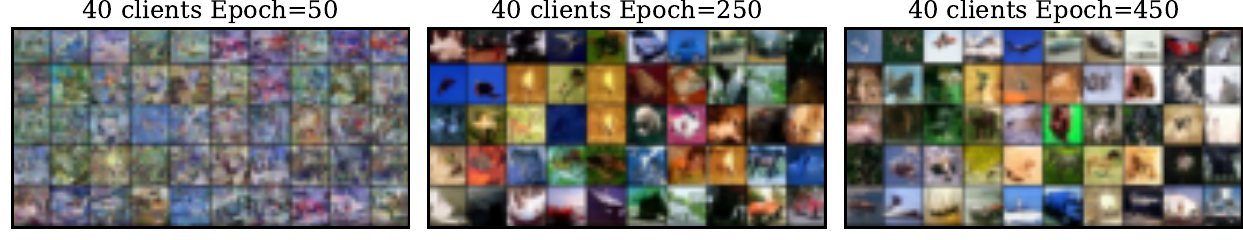}
    \caption{Example simulated images from different training epochs on CIFAR10. The number of training images per client is 875. When the number of clients is high, the longer we train the DDPM, the better images we can obtain.}
    \label{fig:example_synthetic_images_cifar10}
\end{figure}

\subsubsection{Differentially private synthetic images}

\begin{figure}[ht!]
    \centering
    \includegraphics{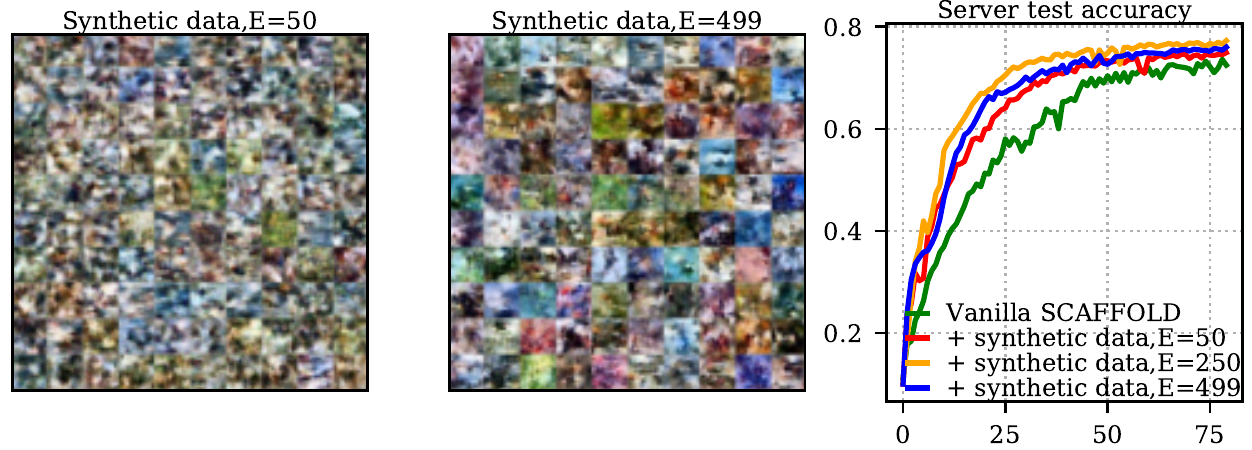}
    \caption{Example of the differentially private synthetic images from when the DP-DDPM has been trained for 50 epochs (\texttt{E=50}) and for 499 epochs (\texttt{E=499}), with respectively. The performance of Fedssyn using SCAFFOLD as the baseline is shown on the right. We achieve better performance and converge slightly faster than the vanilla SCAFFOLD even if the synthetic images are privacy friendly.}
    \label{fig:scaffold_dp_performance}
\end{figure}

\subsubsection{Convergence parameters in DNN experiments}\label{appendix_sec:convergence_param_in_dnn}
\begin{figure}[ht!]
    \centering    \includegraphics{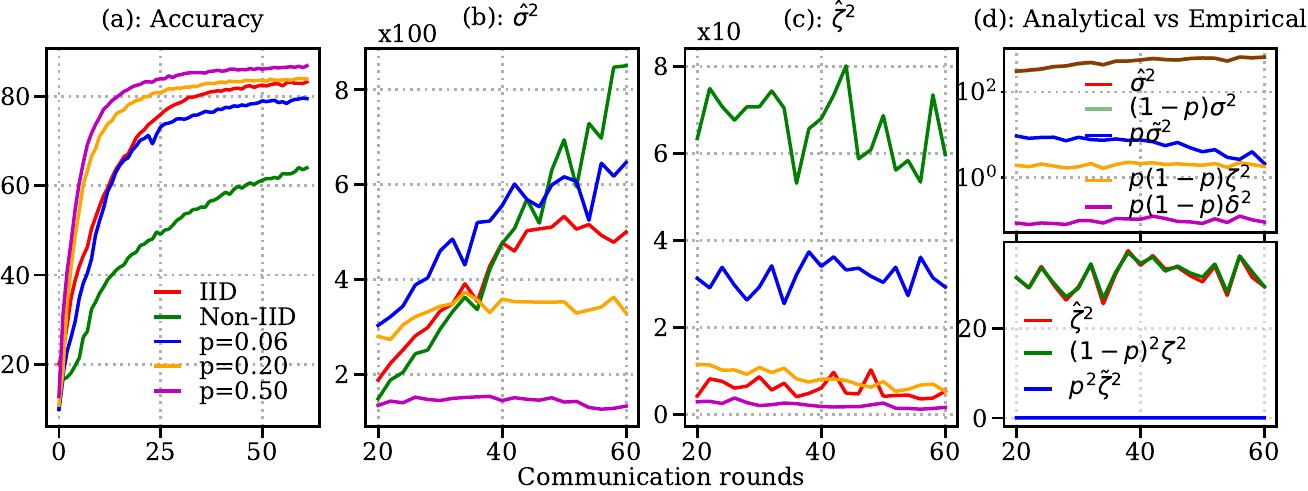}
    \caption{Influence of using shuffled synthetic dataset on the stochastic noise $\hat{\sigma}^2$ and function dissimilarity $\hat{\zeta}^2$ (CIFAR10, 10 clients, $\alpha=0.1$). The percentage of shuffled data in $(d)$ is 0.06. The empirical observation of $\hat{\sigma}^2$ and $\hat{\zeta}^2$ matches the theoretical statements.}
    \label{fig:influence_of_sigma_zeta}
\end{figure}

We here investigate the impact of using shuffled synthetic data on the parameters in the convergence rate for DNN-based FedAvg in Fig.~\ref{fig:influence_of_sigma_zeta}. We use CIFAR10, 10 clients with full participation, $\alpha=0.1$, and $\rho\cdot n_i=4500$. We observe that when $p=0.06$, though the stochastic noise $\hat{\sigma}_p^2$ remains similar, $\hat{\zeta}_p^2$ has reduced by half and we improve the Top-1 accuracy by 20\%. When $p=0.5$, we obtain a much smaller $\hat{\sigma}_p^2$ and $\hat{\zeta}_p^2$. Consequently, we achieve an even better Top-1 accuracy than the IID experiment. However, the parameters in Fig.~\ref{fig:influence_of_sigma_zeta} (b) and (c) are evaluated with different server models $\xx_{p,r}$, so it is less comparable to Lemma 1 where it requires the upper bound and the same server model. Therefore, we evaluate the gradient dissimilarity $\hat{\zeta}_p^2$ and stochastic noise $\hat{\sigma}_p^2$ using $\mathcal{D}_i\cup\tilde{\mathcal{D}}_{si}$ and the corresponding $\sigma^2$ and $\zeta^2$ using $\mathcal{D}_i$ with the same server model $\xx_r$ in each round in Fig.~\ref{fig:influence_of_sigma_zeta} (d). We observe that $(1-p)\sigma^2$ dominates over other terms in the effective stochastic noise, which means the first term in the convergence rate for non-convex function in Corollary I can be simplified as {\footnotesize $\mathcal{O}\left(\frac{(1-p)\sigma_p^2L_p}{n\epsilon^2} \right)$} in this experimental setup. For $\hat{\zeta}_p^2$, the empirical result also matches the theory. These results show that in this experimental setup, adding shuffled synthetic dataset reduces both stochastic noise and function dissimilarity, and lead to a greater accuracy improvement.%

\subsubsection{Influence of the $\delta$ parameter in DNN experiments}
\begin{figure}[ht!]
    \centering    \includegraphics{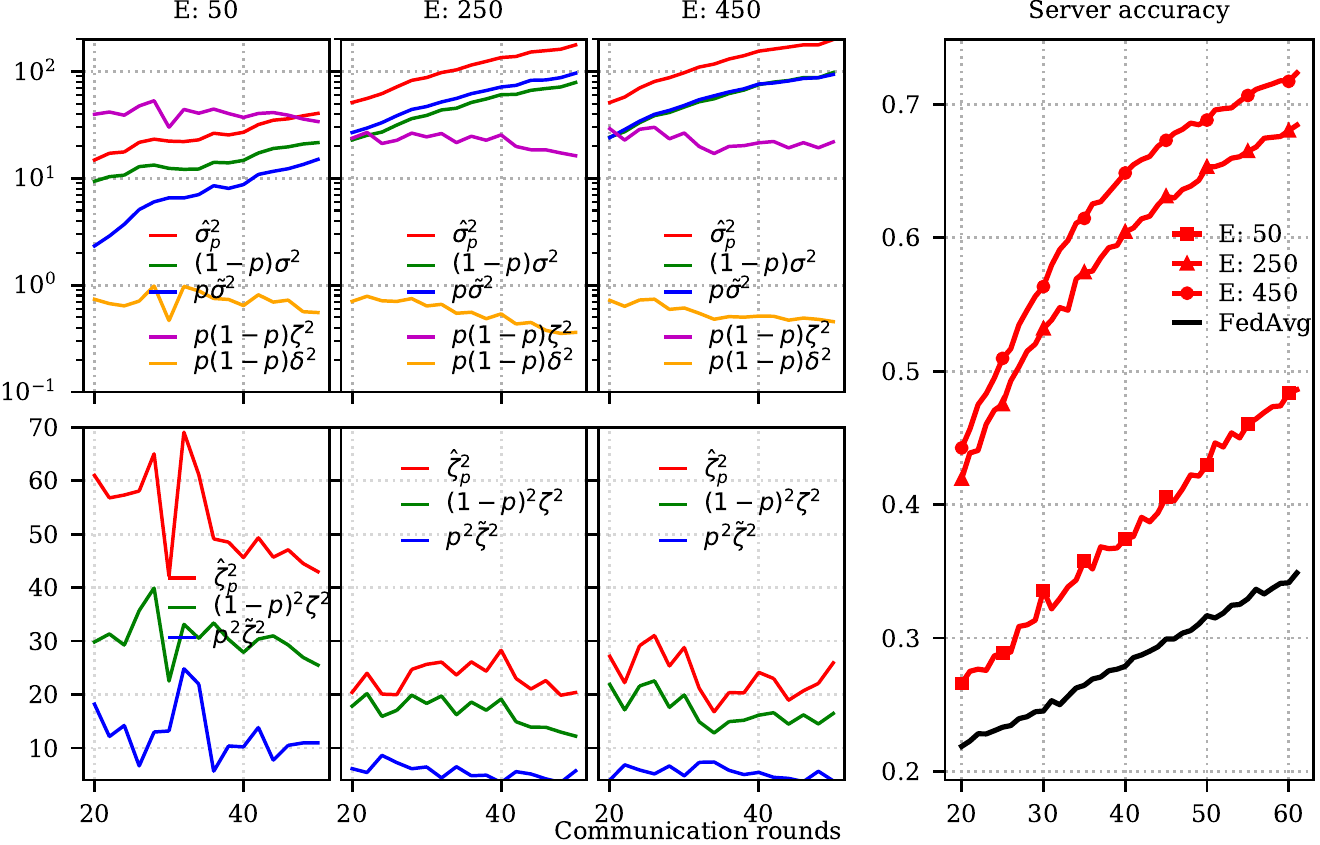}
    \caption{The movement of the convergence parameters given synthetic images that are extracted from different stages of training, e.g., $\texttt{E: 50}$ means the synthetic data shuffled to each client in Fedssyn is generated with the DDPM that has been trained for 50 epochs. We use CIFAR10, 40 clients, $\alpha=0.01, p=0.57$ here.}
    \label{fig:influence_of_delta_parameters}
\end{figure}

We show the impact of the $\delta$ parameter ($||\nabla f(\xx, \mathcal{D}) - \nabla f(\xx, \tilde{\mathcal{D}})||^2\leq\delta^2$) on the convergence parameters $\hat{\sigma}^2$ and $\hat{\zeta}^2$ here. We quantify the values of the stochastic noise and gradient dissimilarity following the procedure described above. In Fig.~\ref{fig:influence_of_delta_parameters}, $\texttt{E: 50}$ means the synthetic data shuffled to each client is generated with the DDPM that has been trained for 50 epochs. Note that Fig.~\ref{fig:influence_of_delta_parameters} only serves as an illustration. We cannot directly match Fig.~\ref{fig:influence_of_delta_parameters} with Lemma 1 as we here show the precise quantified value using different server models along the communication round rather than an upper bound. Under the same server model (i.e., vertical line in each figure), we observe that $\delta$ contributes less to the effective stochastic noise than other parameters such as $\sigma^2$ and $\tilde{\sigma}^2$. However, it is worth remembering that $\delta$ also can influence the effectiveness of the server model (see the server accuracy in Fig.~\ref{fig:influence_of_delta_parameters}) and $\tilde{\mathcal{D}}$ is different across different FL experiments (i.e., from \texttt{E:50} to \texttt{E:450}), so we cannot compare the curves horizontally. 

\subsection{Experimental results on dSprites dataset}\label{appendix_sec:dsprite}
In this section, we demonstrate the effectiveness of our proposed framework on the dSprites dataset~\cite{dsprites17}. We here perform a slightly different synthetic data generation process. Rather than using the DDPM~\cite{ho2020denoising}, we generate the synthetic dataset on each client with $\beta$-VAE~\cite{https://doi.org/10.48550/arxiv.1804.03599}. We then aggregate and shuffle the collected synthetic dataset on the server and distribute them to clients following the same procedure as described in Sec.~\textcolor{red}{3} and Sec.~\textcolor{red}{4}.

\subsubsection{Training details}
The dSprites dataset~\cite{dsprites17} contains three different types of shapes with different colours, scales, rotations, and locations. 

\begin{figure}[ht!]
    \centering
    \includegraphics[width=.8\textwidth]{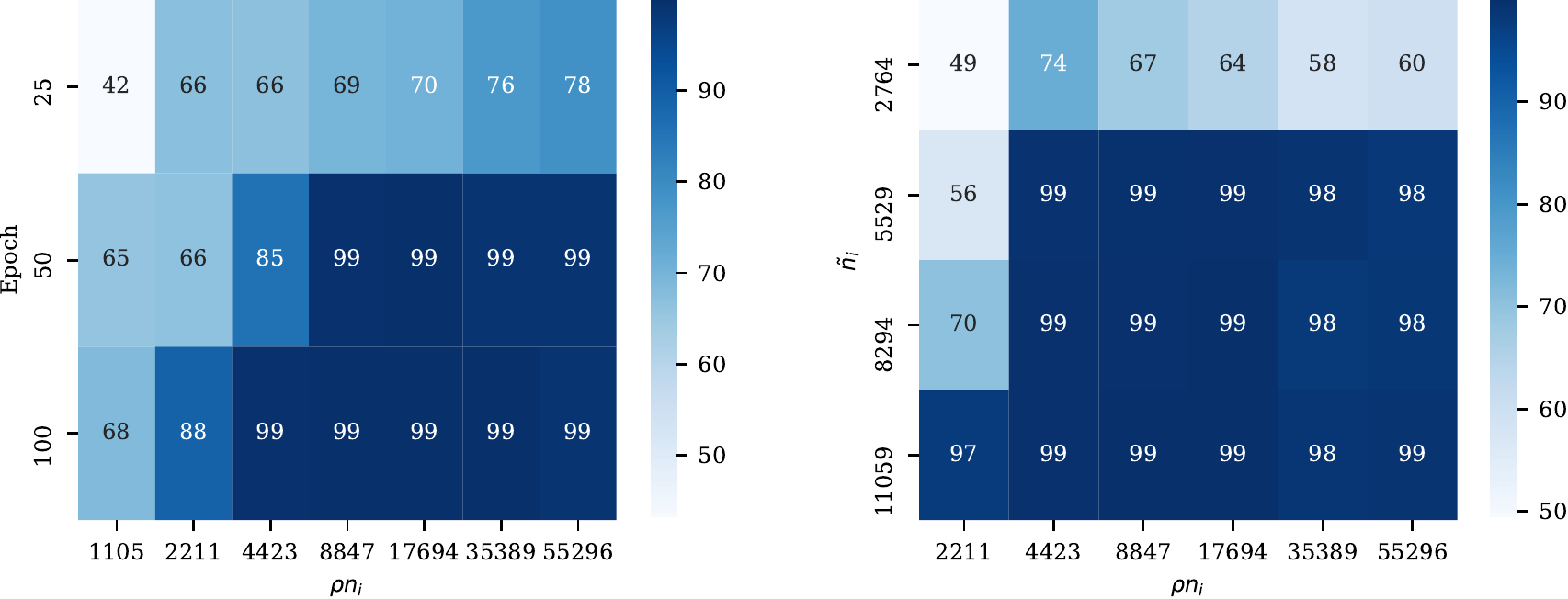}
    \caption{Performance on the dSprites test dataset. Left: the influence of the number of training epochs and training images ($\rho n_i$) in training the generator on the FL performance. Right: the influence of the number of synthetic images ($\tilde{n}_i$) on the FL performance. When we use fewer training images to train the generator (e.g. 2211), it is beneficial to train the generator longer and sample more synthetic images.}
    \label{fig:dsprite_performance}
\end{figure}

\textbf{Data preparation}  We first randomly select 10\% of the images from the dSprites dataset to formulate the test set. We leave the test set on the server to evaluate the performance of the server model. For the rest of the dataset, we split them among 12 clients ($N=12$) based on the four spatial locations (top-left, bottom-left, top-right, or bottom-right) and three shapes (square, ellipse, or heart) such that each client only sees a single type of shape from one of the four pre-defined locations. The number of images on each client is the same ($n_i=55296$). We consider each shape as a class and perform shape classification. 

\begin{figure}[ht!]
    \centering
    \includegraphics[scale=0.5]{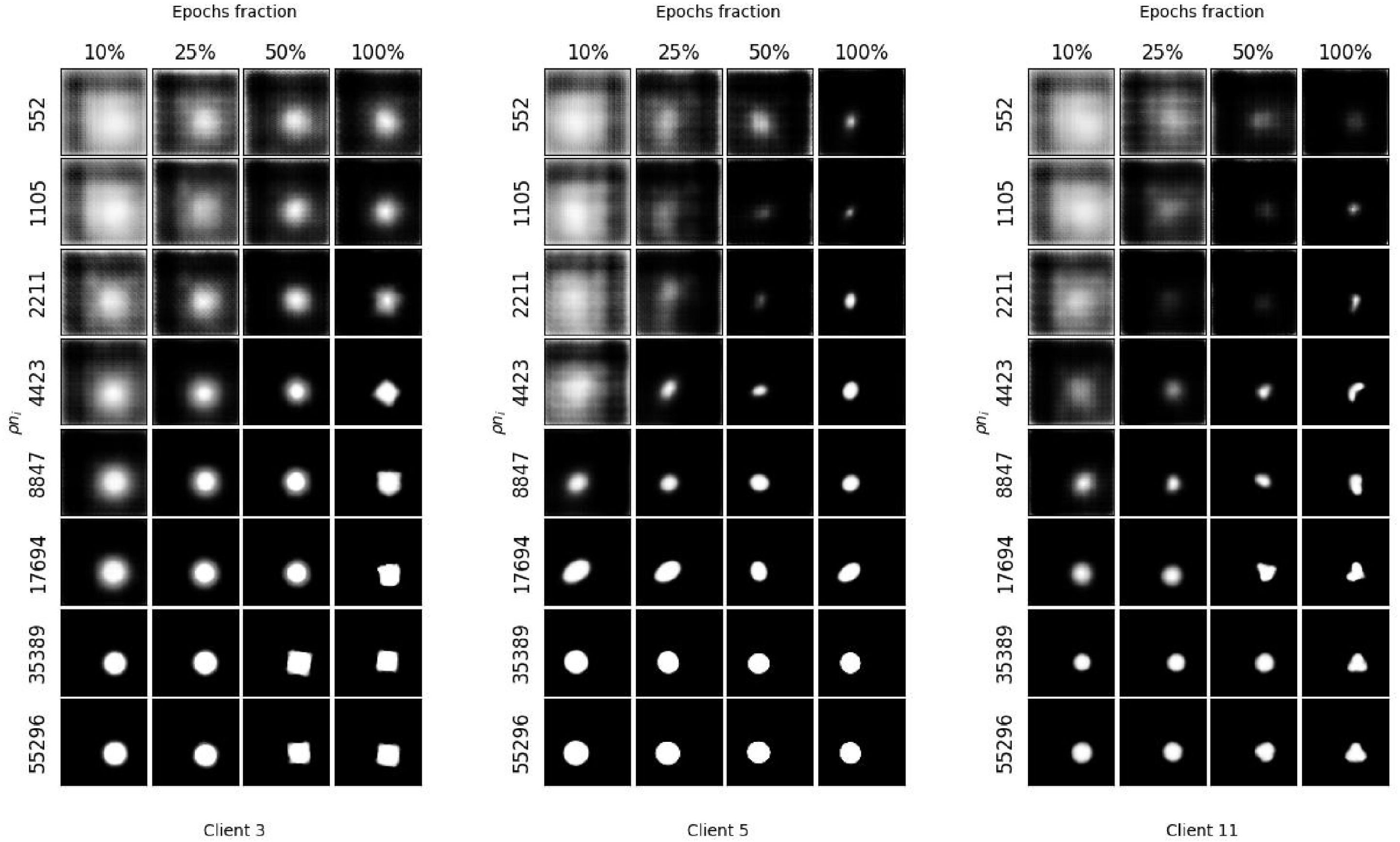}
    \caption{Example synthetic images from clients 3, 5, and 11. The x-axis shows the training fraction of 360 epochs, and the y-axis shows the number of training images used to train the generator. The number of training images has a more substantial influence on the quality of the synthetic data.} 
    \label{fig:dSprites_visualizations_from_checkpoints}
\end{figure}
\noindent
\textbf{Synthetic data generation} 
We use $\beta$-VAE with the same architecture as~\cite{https://doi.org/10.48550/arxiv.1804.03599} following a publicly available implementation
We train an individual $\beta$-VAE~\cite{https://doi.org/10.48550/arxiv.1804.03599} on each client with batch size 256, latent dimension 10. To evaluate the sensitivity of the federated learning optimization on the quality of the generator, we train the generator using different fractions of data $\rho n_i$ with $\rho$ being $\{1\%, 2\%, 4\%, 8\%, 16\%, 32\%, 64\%, 100\% \}$ on each client. When $\rho=100\%$, we train a $\beta$-VAE with the entire dataset from each client. 


Many works have demonstrated that the latent variables in $\beta$-VAE can encode disentangled representative features of the input images~\cite{https://doi.org/10.48550/arxiv.1804.03599, dsprites17,Kingma2013AutoEncodingVB}, and manipulating a single latent dimension can result in substantial changes of the corresponding factor of variation, e.g., scale, in the output from the decoder. Therefore, we can obtain diverse synthetic images by interpolating the extracted latent representations. To achieve this, we extract the averaged latent representations for each class, specifically the mean of the posterior distribution $\mu_c$~\cite{https://doi.org/10.48550/arxiv.1804.03599}. We then sample $\tilde{n}_i$ latent codes from the Gaussian distribution parameterised by mean $\mu_c$ and standard deviation $1$. The sampled latent codes are then passed as the input for the decoder from $\beta$-VAE to generate synthetic images. We collect the synthetic images from all the clients, which are then shuffled and distributed equally to each client such that the local dataset on each client is $p\mathcal{D}_i + (1-p)\tilde{\mathcal{D}}_{si}$. An example of the generated synthetic image is shown in Fig.~\ref{fig:dSprites_visualizations_from_checkpoints}. With the updated dataset on each client, we follow the same procedure as documented in Sec.~\textcolor{red}{4} to perform federated optimization with FedAvg~\cite{McMahan2016CommunicationEfficientLO}. 



\subsubsection{Discussion}

We show the sensitivity of FedAvg's performance on the quality of the synthetic images in Fig.~\ref{fig:dsprite_performance}. The left image shows that training the VAE longer does not necessarily provide better quality synthetic images when we use fewer training images. However, when we use more training images (higher $\rho$), the images extracted from early checkpoints, e.g., 50\% epochs, are already of high quality. We observe a similar pattern in the right image in Fig.~\ref{fig:dsprite_performance}, where the number of training images has a more substantial influence on the performance of FedAvg than the number of generated synthetic images.

\end{document}